\documentclass[runningheads]{llncs}

\usepackage{eccv}

\usepackage{eccvabbrv}

\usepackage{graphicx}
\usepackage{booktabs}

\usepackage[accsupp]{axessibility}  

\usepackage[pagebackref,breaklinks,colorlinks,citecolor=eccvblue]{hyperref}

\usepackage{orcidlink}

\usepackage{mmstyles}
\usepackage{multirow}
\usepackage[ruled]{algorithm2e}

\usepackage{amsmath,amsfonts,bm}

\def\eqref#1{equation~(\ref{#1})}

\def\1{\bm{1}}

\def\vmu{{\bm{\mu}}}

\def\va{{\bm{a}}}

\def\vd{{\bm{d}}}

\def\vf{{\bm{f}}}

\def\vh{{\bm{h}}}

\def\vm{{\bm{m}}}

\def\vv{{\bm{v}}}

\def\vx{{\bm{x}}}
\def\vy{{\bm{y}}}

\def\mI{{\bm{I}}}

\def\mM{{\bm{M}}}

\DeclareMathAlphabet{\mathsfit}{\encodingdefault}{\sfdefault}{m}{sl}
\SetMathAlphabet{\mathsfit}{bold}{\encodingdefault}{\sfdefault}{bx}{n}

\def\gN{{\mathcal{N}}}

\begin{document}

\title{GetMesh: A Controllable Model for High-quality Mesh Generation and Manipulation} 

\titlerunning{GetMesh: High-quality Mesh Generation and Manipulation}

\newcommand{\AuthorSpace}{\hspace{1.2em}}
\author{%
Zhaoyang Lyu$^{1}$\thanks{Equal Contribution.} \AuthorSpace{} Ben Fei$^{1,2 \star}$ \AuthorSpace{} Jinyi Wang$^{3\star}$\AuthorSpace{} Xudong Xu$^{1}$\AuthorSpace{} \\
Ya Zhang$^{1,3}$\AuthorSpace{} Weidong Yang$^{2}$\AuthorSpace{} Bo Dai$^{1}$ 
}


\authorrunning{Z.~Lyu et al.}

\institute{Shanghai Artificial Intelligence Laboratory \and Fudan University \and
Shanghai Jiao Tong University}

\maketitle

\newcommand{\name}{\texttt{GetMesh}}
\newcommand{\reorganize}{\text{re-organize}}
\newcommand{\appd}{\textcolor{red}{Appendix}}
\newcommand{\second}[1]{#1}

\begin{abstract}
\vspace{-0.5em}
Mesh is a fundamental representation of 3D assets in various industrial applications, and is widely supported by professional softwares. 
However,
due to its irregular structure,
mesh creation and manipulation is often time-consuming and labor-intensive. 
In this paper, 
we propose a highly controllable generative model, \name, for
mesh generation and manipulation across different categories.
By taking a varying number of points as the latent representation, and re-organizing them as triplane representation,
\name~generates meshes with rich and sharp details, outperforming both single-category and multi-category counterparts.
Moreover,
it also enables fine-grained control over the generation process that previous mesh generative models cannot achieve,
where changing global/local mesh topologies, adding/removing mesh parts, and combining mesh parts across categories can be intuitively, efficiently, and robustly accomplished by adjusting the number, positions or features of latent points. 
Project page is \url{https://getmesh.github.io}.
\end{abstract}
\section{Introduction}
\label{sec:intro}

\begin{figure*}[h]
\centering
\vspace{-1em}
\includegraphics[width=1\linewidth]{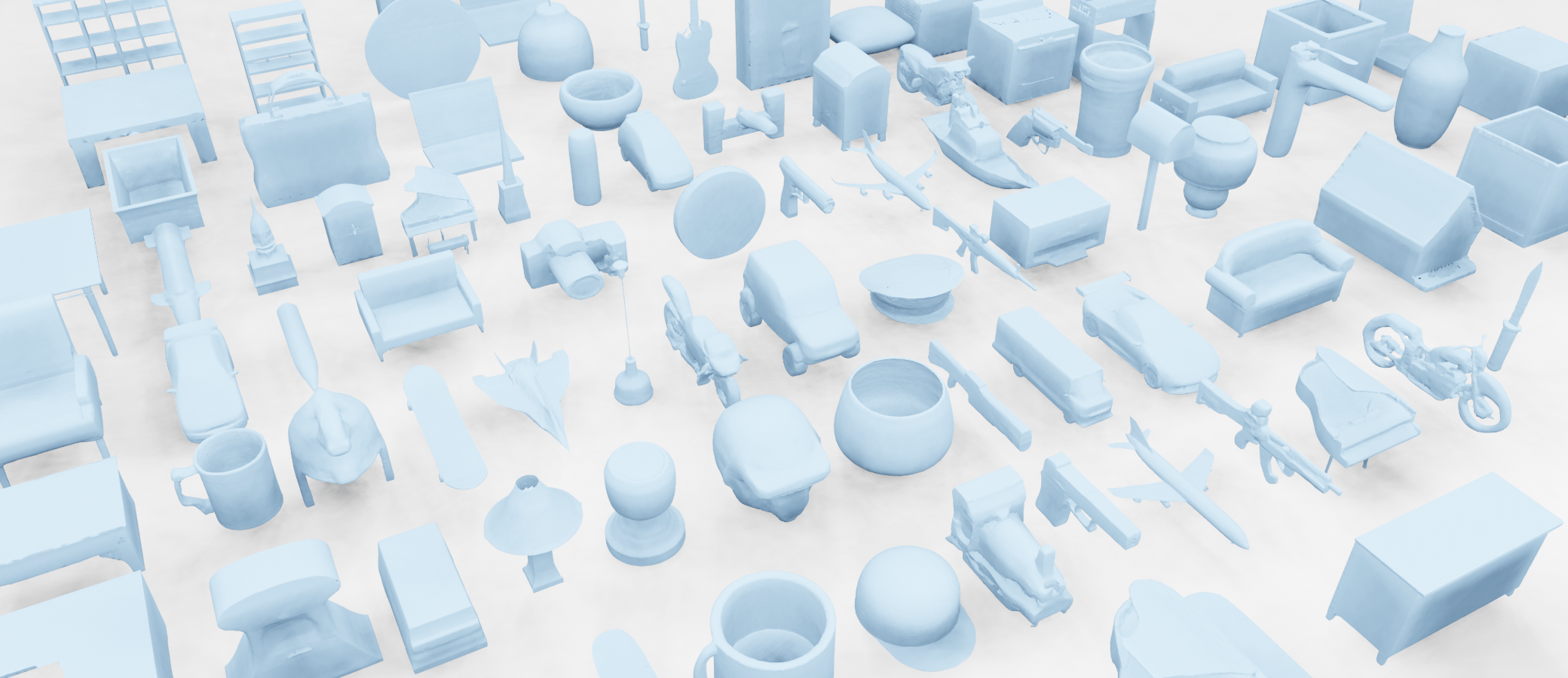}
\vspace{-1em}
\caption{Meshes generated by our method. GetMesh is able to generate diverse and high-quality meshes across the $55$ categories in ShapeNet.}
\label{fig: teaser_no_texture}
\vspace{-1.5em}
\end{figure*}

3D asset generation is of significant value for AR/VR, gaming, filming and design.
Meshes are a kind of fundamental representation for 3D assets in industrial applications since their creation, editing, and rendering are supported by many professional softwares.
However, mesh creation and editing can be quite time-consuming and require extensive artistic training,
since meshes are irregular in their data structure with varying topologies across different instances and categories.
Therefore, a controllable paradigm that supports intuitive and efficient generation and manipulation of meshes is of great need.

While directly operating on meshes is difficult and sometimes infeasible,
an alternative is to rely on a suitable latent representation and associate it with mesh via a pair of encoder and decoder.
Researchers have designed many types of latent representations for mesh generation. 
Among them, point-based representations~\cite{Zeng2022LIONLP, lyu2023controllable} are compact and convenient for editing, but rely on a Poisson reconstruction-like algorithm~\cite{Peng2021ShapeAP} to obtain meshes, which are overly smooth without sharp details.
Voxel-based representations~\cite{Li2022DiffusionSDFTV, Liu2023MeshDiffusionSG, Zheng2023LocallyAS, Zheng2022SDFStyleGANIS} are proposed for 3D shapes due to their regular data structure,
yet their computational and memory cost increase significantly for high-quality meshes.
Finally, triplane-based representations~\cite{Gao2022GET3DAG, gupta20233dgen, Shue20223DNF} are compact and efficient for high-quality mesh modeling, but can be quite obscure and inefficient to control as they are squeezed along a specific dimension.
Therefore,
despite the blooming attention and progress in text-to-3D \cite{Poole2022DreamFusionTU, Lin2022Magic3DHT, Xu2022Dream3DZT, Chen2023Fantasia3DDG, Wang2023ProlificDreamerHA} generation,
finding a suitable latent representation and subsequently an intuitive and efficient paradigm for controllable mesh generation and manipulation remains an open question.

In this paper, we propose a novel model, \name, a multi-category generative model that enables both high-quality mesh generation and flexible control over the generation process.
It combines the merits of both point-based representation and triplane-based representation,
which is achieved by using \emph{a varying number of points} as the latent representation, 
and \emph{re-organizing them as triplane} representation.
The feature and position distributions of these points can be respectively modeled by two diffusion models.
A varying number of points as the latent representation provides significant controllability over the generation process,
where changing global/local topologies of meshes, adding/removing mesh parts, as well as combining mesh parts across instances/categories can all be achieved intuitively, efficiently, and robustly by adjusting the number, positions, or features of these points.
To avoid obtaining overly smoothed meshes from point-based representation,
the proposed paradigm re-organizes these points by projecting their features onto \emph{xy,xz,yz} planes according to their positions,
forming a triplane-based representation.
Subsequently,
a triplane-based decoder with a refinement module can thus be used to extract high-quality meshes with sharp details.

We conduct extensive experiments on ShapeNet~\cite{chang2015shapenet} to evaluate \name.
As expected, 
\name~generates meshes with rich and sharp details as shown in Figure~\ref{fig: teaser_no_texture},
outperforming its multi-category counterparts,
and even single-category generative models.
Moreover,
since \name~is capable of controlling mesh generation intuitively and flexibly,
\name~is able to change the topology of generated meshes such as turning a twin-engine airplane into a four-engine one, 
and gradually turning a car into an airplane as shown in Section~\ref{sec: controllable_mesh_generation}.
\name~also successfully combines mesh parts across different categories,
leading to a car with airplane wings, and a table with lamp top.
Finally,
\name~is shown to work well with an off-the-shelf material generative method to acquire materials for its generated meshes.

\section{Background}
\label{sec:background}

\begin{figure*}[tb]
\vspace{-1em}
\centering
\includegraphics[width=1\linewidth]{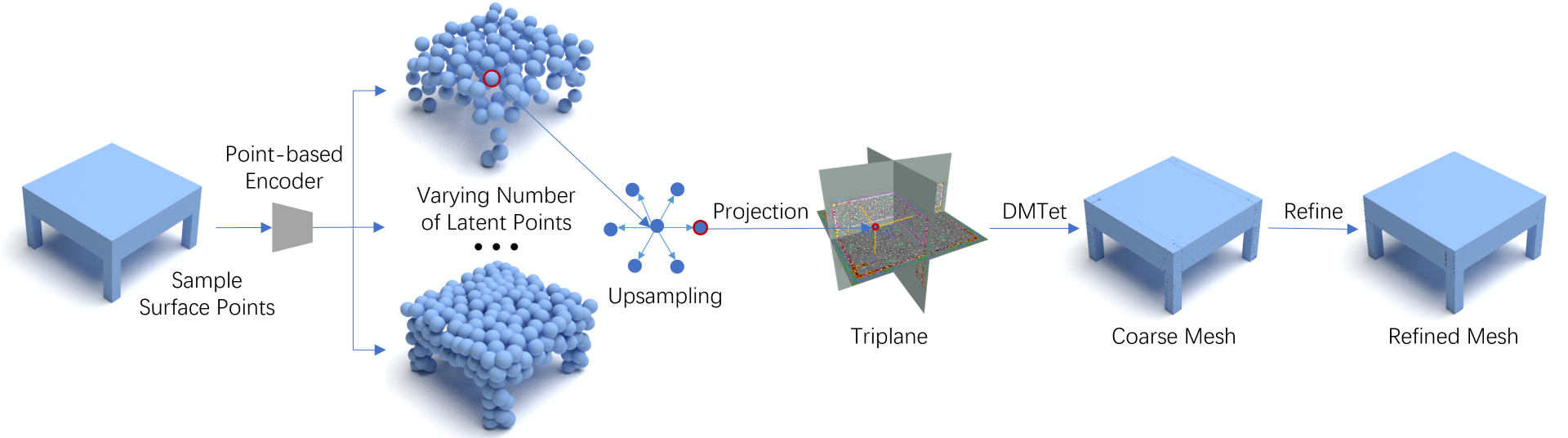}
\vspace{-1.5em}
\caption{Overview of the mesh autoencoder.
Points are sampled from the surface of the input mesh and encoded to a varying number of latent points.
The latent point representation is re-organized to the triplane representation by projecting the points to the triplane.
DMTet is utilized to extract a coarse mesh from the triplane and a refinement module further refines the coarse mesh. 
}
\label{fig: mesh_ae_overview}
\vspace{-1.5em}
\end{figure*}

Denoising Diffusion Probabilistic Models (DDPMs) are generative models that learn the distribution of samples in a dataset.
A DDPM is composed of two processes: the diffusion process and the reverse process.
The diffusion process gradually adds noise to clean samples $\vx^0$ and turns them into Gaussian noises $\vx^T$ after $T$ steps. It is defined as 
\vspace{-0.5em}
\begin{align}
\label{eqn:diffusion_process}
    q(\vx^1,\cdots,\vx^T|\vx^0) = \prod_{t=1}^T q(\vx^t|\vx^{t-1}), \\
    \text{ where }
    q(\vx^t|\vx^{t-1})=\gN(\vx^t;\sqrt{1-\beta_t}\vx^{t-1},\beta_t \mI),
\end{align}
$\gN$ is the Gaussian distribution.
In our experiments, we set $T=1000$, and $\beta_t$ linearly increase from $1\times10^{-4}$ to $2\times10^{-2}$ as $t$ increases from $1$ to $T$.
The reverse process is the data generation process.
It starts from a Gaussian noise $\vx^T$ and denoises it step by step, eventually turning it into a clean sample $\vx^0$.
The reverse process is formally defined as
\vspace{-0.5em}
\begin{align}
\label{eqn:reverse_process}
\begin{split}
    p_{\bm{\theta}}(\vx^0,\cdots,\vx^{T-1}|\vx^T)=\prod_{t=1}^T p_{\bm{\theta}}(\vx^{t-1}|\vx^t), \\
    \text{ where }
    p_{\bm{\theta}}(\vx^{t-1}|\vx^t) = \gN(\vx^{t-1};\bm{\mu}_{\bm{\theta}}(\vx^t, t), \tilde{\beta}_t \mI),
\end{split}
\end{align}
$\tilde{\beta}_t = \frac{ 1-\bar{\alpha}_{t-1} }{ 1-\bar{\alpha}_{t} } \beta_t.$
We follow \cite{ho2020denoising} to reparameterize the mean $\bm{\mu}_{\bm{\theta}}(\vx^t, t)$ as 
\vspace{-0.2em}
\begin{align}
\bm{\mu}_{\bm{\theta}}(\vx^t, t) = \frac{1}{\sqrt{\alpha_t}}\left(\vx^t-\frac{\beta_t}{\sqrt{1-\bar{\alpha}_t}}\bm{\epsilon}_{\bm{\theta}}(\vx^t, t)\right),
\end{align}
where $\alpha_t = 1 - \beta_t$, $\bar{\alpha}_t = \prod_{i=1}^t\alpha_i$, and $\bm{\epsilon}_{\bm{\theta}}$ is a neural network with parameters specified by $\bm{\theta}$.
The loss to train the network is
\begin{align}
\label{eqn: training objective}
    L(\bm{\theta}) = 
    \mathbb{E}_{\vx^0 \sim p_{\text{data}}}\ 
    \|\bm{\epsilon} - \bm{\epsilon}_{\bm{\theta}}(\sqrt{\bar{\alpha}_t}\vx^0 + \sqrt{1-\bar{\alpha}_t}\bm{\epsilon}, t)\|^2,
\end{align}
where $p_{\text{data}}$ is the distribution of the dataset, $t$ is sampled uniformly between $1$ and $T$, and $\bm{\epsilon}$ is a Gaussian noise.

\section{Methodology}
\label{sec:method}

A mesh is represented as vertices and faces, where faces describe the connections between vertices.
It is difficult to directly train generative models on meshes due to their irregular data structure and discrete connections between vertices.
Therefore, we first train an autoencoder to encode a mesh to a point-based representation that is easier to process for neural networks.
Then we train DDPMs in the latent space of the autoencoder with the same merit as \cite{vahdat2021score,rombach2022high},
and high-quality meshes can be reconstructed from the generated latent representation by a decoder.

We first describe the architecture of the mesh autoencoder.
Its overall architecture is shown in Figure~\ref{fig: mesh_ae_overview}.
We sample points from the input mesh and use a point-based encoder to encode the points to a varying number of latent points and their features.
Then we re-organize the latent point representation to a triplane representation.
Finally, we use a triplane-based decoder with a refinement module to decode the triplane representation to a high-quality mesh.
The detailed architecture of the mesh autoencoder is explained in the following sections.

\subsection{Point-based Encoder}
We sample a point cloud $\va \in \mathbb{R}^{N_{\text{in}} \times 3}$ from the surface of the input mesh and the encoder encodes the sampled point cloud to a set of latent points $\vx \in \mathbb{R}^{N \times 3}$ with features $\vy \in \mathbb{R}^{N \times D}$, where $N_{\text{in}}$ is the number of input points,  $N$ is the number of latent points and $D$ is the feature dimension.
To enable more versatile editing like addition and deletion of latent points, we design an encoder that supports a varying number of latent points $\vx$.
The latent points are sampled using Furthest Point Sampling (FPS) from the input point cloud $\va$, and the number of latent points $N$ is uniformly sampled in the interval $[N_{\text{min}}, N_{\text{max}}]$ during training, where $N_{\text{min}}, N_{\text{max}}$ are two hyperparameters that control the minimum and maximum number of latent points.
The architecture of the encoder resembles the one proposed in \cite{lyu2023controllable}, which is an improved version of PointNet++~\cite{qi2017pointnet++} proposed in \cite{lyu2021conditional}.
The encoder gradually downsamples the input point cloud and propagates features level by level, until features are propagated to the sampled latent points.
We refer readers to the original paper~\cite{lyu2023controllable} for details of the encoder and Appendix Section~\ref{sec: appendix_mesh_autoencoder_architecture} on how we encode the input point cloud to a varying number of latent points.

\subsection{Re-organize Latent Points as Triplane}
In the previous section, we use a point-based encoder to encode the input mesh to a set of latent points and features.
While this latent point representation is compact and intuitive to edit, it is difficult to reconstruct high-quality meshes with sharp details and thin structures directly from point-based representation using Poisson reconstruction or other related algorithms.
In light of the recent success of triplane representation~\cite{Gao2022GET3DAG, gupta20233dgen, Shue20223DNF} in reconstructing relative high-quality meshes, we propose to re-organize the latent point representation as a triplane representation.
The basic idea is to project the latent points together with their features onto the triplane.
Specifically, for each point, we project it to the three perpendicular planes according to its position. Its feature is fed to a shared MLP and the output feature is assigned to the corresponding three pixels.
Features of latent points projected to the same pixel are aggregated by average pooling.
Pixels that do not correspond to any latent points are filled with zeros.
In the point-based encoder, we set the number of latent points small in order to make the latent space compact.
To make the triplane representation concrete, we first upsample the latent points $\vx$ to a denser point cloud 
$\vx_{\text{u}}$
, and propagate features from $\vx$ to $\vx_{\text{u}}$. 
Then we project the upsampled point cloud $\vx_{\text{u}}$ to the triplane.
More details of the upsampling process are provided in Appendix Section~\ref{sec: appendix_mesh_autoencoder_architecture}.

\begin{figure}[t]
\centering
\vspace{-1em}
\includegraphics[width=0.6\linewidth]{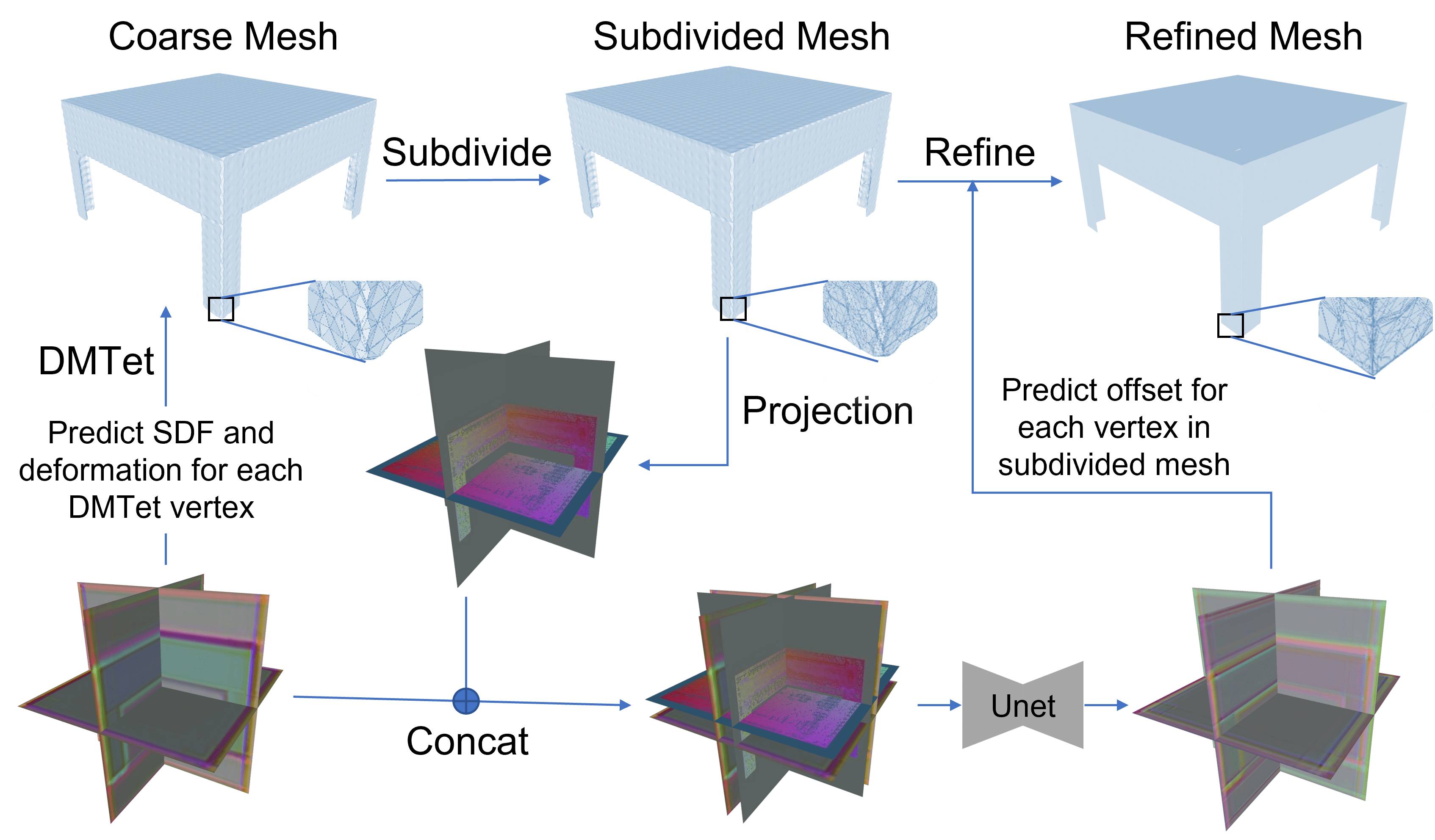}
\vspace{-1em}
\caption{Architecture of the refinement module.}
\label{fig: mesh_refine_module}
\vspace{-1.5em}
\end{figure}

\subsection{Triplane-based Decoder}
After re-organizing the latent point representation as the triplane representation, we use a triplane-based decoder together with a refinement module to reconstruct a high-quality mesh from the triplane representation.
The architecture of the decoder is similar to ~\cite{gupta20233dgen}.
The triplane is first processed by a UNet with 3D aware convolutions proposed in~\cite{wang2023rodin} and we obtain the processed triplane feature $\vf \in \mathbb{R}^{3 \times H \times W \times C}$.
Next, DMTet~\cite{shen2021deep} is utilized to extract a mesh differentiably.
The SDF value and deformation of each vertex in the DMTet grid are predicted by querying the triplane feature $\vf$.
Specifically, for each vertex $\vv$ in the DMTet grid, it is projected onto the triplane and we obtain three features $\vf_{\vv}^{\text{xy}}, \vf_{\vv}^{\text{yz}}, \vf_{\vv}^{\text{zx}}$ through bilinear interpolation of the feature planes.
The three features $\vf_{\vv}^{\text{xy}}, \vf_{\vv}^{\text{yz}}, \vf_{\vv}^{\text{zx}}$ are concatenated and fed to two MLPs to predict the SDF value and deformation of the vertex $\vv$, respectively.
Next, a mesh can be extracted from the DMTet grid using the differentiable Marching Tetrahedra algorithm.

\noindent \textbf{Refinement.}
We find that meshes extracted from DMTet often bear artifacts as shown in Figure~\ref{fig: mesh_refine_module}.
The edge of the extracted mesh is not sharp, and the surface is not smooth: There are evenly distributed tetrahedron-shaped dimples and bumps on the mesh surface.
More examples are provided in Figure~\ref{fig: compare_coarse_and_refined}.
We also observe similar artifacts in meshes generated by Get3D~\cite{Gao2022GET3DAG}, which uses DMTet to extract meshes as well.
It is probably because of the low resolution of the DMTet grid adopted ($128^3$), and the underlying tetrahedral representation of the DMTet grid.
We could try to increase the resolution of the DMTet grid, but the memory and computational cost will increase dramatically, and the topology of the tetrahedral grid could still affect the extracted mesh.
Instead, we subdivide and refine the coarse mesh $\mM$ extracted from DMTet.
The architecture of this refinement module is shown in Figure~\ref{fig: mesh_refine_module}.
We subdivide each face in the extracted mesh by adding new vertices at the middle point of each edge and connecting them to form $4$ smaller faces.
The 3D coordinates of vertices of the new mesh $\mM'$ are fed to a shared MLP to extract features and then projected to a triplane feature $\vh \in \mathbb{R}^{3 \times H \times W \times C}$.
We concatenate $\vh$ with $\vf$ along the channel dimension and obtain a new triplane feature $[\vf, \vh] \in \mathbb{R}^{3 \times H \times W \times 2C}$.
The concatenated triplane feature $[\vf, \vh]$ is fed to a lightweight UNet and we obtain the processed triplane feature $\vh' \in \mathbb{R}^{3 \times H \times W \times C'}$.
We refine the subdivided mesh $\mM'$ by predicting a displacement for each vertex in $\mM'$ using the triplane feature $\vh'$.
The method is the same as the one that uses $\vf$ to predict the displacement of each vertex in the DMTet grid.
More details of the decoder and refinement module are provided in Appendix Section~\ref{sec: appendix_mesh_autoencoder_architecture}

\noindent \textbf{Training Loss.} The supervision of the autoencoder is added on the latent feature $\vy$, the upsampled points $\vx_{\text{u}}$, and the reconstructed mesh $\mM'$.
For $\vy$, we add a Kullback-Leibler divergence loss between $\vy$ and the standard Gaussian distribution with weight $10^{-7}$ in order to make the distribution of latent features relatively simple and smooth.
For $\vx_{\text{u}}$, we first downsample the input point cloud $\va$ to the same number of points as $\vx_{\text{u}}$ using FPS, and then add a Chamfer distance (CD) between the downsampled points and $\vx_{\text{u}}$ with weight $1$.
For $\mM'$, we add a rendering-based loss similar to the one used in~\cite{gupta20233dgen}.
Specifically, $\mM'$ is fed to a differentiable renderer to obtain the mask silhouette $\vm$ and depth map $\vd$, and the rendering-based loss is computed as the sum of L2 distance between $\vm$ and ground-truth mask silhouette, and L1 distance between $\vd$ and ground-truth depth map, averaged across $N_{\text{view}}$ views.
We also warm up the training of the autoencoder by directly supervising the predicted SDF values of the DMTet vertices.
See more details in Appendix Section~\ref{sec: appendix_mesh_autoencoder_architecture}.

\subsection{Latent Diffusion Models}
\label{sec: latent_diffusion_model}
After training the autoencoder, we train latent diffusion models on the latent point representation.
The representation consists of latent points $\vx \in \mathbb{R}^{N \times 3}$ and features $\vy \in \mathbb{R}^{N \times D}$.
Note that the number of latent points $N$ could vary between $N_{\text{min}}$ and $N_{\text{max}}$.
Similar to~\cite{lyu2023controllable}, we train two DDPMs to model the distribution of $\vx$ and $\vy$, respectively.
The first DDPM $\vepsilon_{\text{position}}$ is named position DDPM and learns the distribution of $\vx$, trained with loss defined in Equation~\ref{eqn: training objective}.
The second DDPM $\vepsilon_{\text{feature}}$ is named feature DDPM and learns the distribution of $\vy$ conditioned on $\vx$.
We use the Transformer architecture in~\cite{pang2022masked} for both $\vepsilon_{\text{position}}$ and $\vepsilon_{\text{feature}}$.
$\vx$ and $\vy$ are padded to the maximum length $N_{\text{max}}$ during training.
See Appendix Section~\ref{sec: appendix_latent_diffusion_model} for more details of the Transformer architecture and training losses.

\noindent \textbf{Sampling.} To sample from the trained two DDPMs, we first use $\vepsilon_{\text{position}}$ to generate positions $\vx$ of the latent points. 
Note that the number of latent points can be chosen arbitrarily between $N_{\text{min}}$ and $N_{\text{max}}$ by the user during sampling.
Then we use $\vepsilon_{\text{feature}}$ to generate features $\vy$ conditioned on $\vx$.
Finally, the generated latent points $\vx$ with features $\vy$ can be reconstructed to a mesh by the trained autoencoder.
\section{Related Work}
\label{sec:related_work}

\noindent \textbf{Diffusion Models.}
Diffusion models are likelihood-based generative models composed of a diffusion process and a reverse process.
They have been thoroughly explored for image generation~\cite{ho2020denoising, Dhariwal2021DiffusionMB, Nichol2021ImprovedDD, Ho2021CascadedDM} and speech synthesis~\cite{Kong2020DiffWaveAV, Jeong2021DiffTTSAD, Popov2021GradTTSAD}.
To alleviate the slow sampling speed of diffusion models, latent diffusion models~\cite{rombach2022high, vahdat2021score} are proposed to train diffusion models in the latent space of an autoencoder, which encodes data samples to a more compact representation and thus accelerates the training and sampling speed of diffusion models.
Our method is based on latent diffusion models.

\noindent \textbf{Diffusion Model for 3D Shape Generation.}
Diffusion models have also been explored for 3D shape generation.
They are first applied to point cloud generation~\cite{Luo2021DiffusionPM, Zhou20213DSG, Zeng2022LIONLP, lyu2023controllable, Nichol2022PointEAS}.
Some methods~\cite{Zeng2022LIONLP, lyu2023controllable} propose to reconstruct meshes from the generated point clouds through surface reconstruction techniques~\cite{Peng2021ShapeAP}.
Another line of works~\cite{Li2022DiffusionSDFTV, Chou2022DiffusionSDFCG, Shue20223DNF, Liu2023MeshDiffusionSG, Zheng2023LocallyAS, gupta20233dgen, Nam20223DLDMNI} utilizes implicit fields~\cite{Park2019DeepSDFLC, Mescheder2018OccupancyNL} to generate meshes. 
They usually design latent representations such as points, voxels, or triplanes to represent the implicit fields, and then train diffusion models on this latent representation.
Meshes can be later reconstructed by marching cubes~\cite{Lorensen1987MarchingCA} or deep marching tetrahedra (DMTet~\cite{shen2021deep}).

\noindent \textbf{Text-to-Image Model for 3D Shape Generation.}
Recent works~\cite{Poole2022DreamFusionTU, Lin2022Magic3DHT, Xu2022Dream3DZT, Chen2023Fantasia3DDG, Wang2023ProlificDreamerHA} leverage a pre-trained text-to-image diffusion model~\cite{rombach2022high} to generate 3D assets given a text prompt.
They first represent a 3D asset as a neural radiance field (NERF~\cite{ben2021nerf}) or a mesh with a texture map, then render it as images and apply Score Distillation Sampling (SDS) Loss~\cite{Poole2022DreamFusionTU} to optimize the parameters of the 3D asset.
However, these methods are quite time-consuming as they need to optimize the parameters of a 3D asset for every single object.
It is also difficult for text to accurately control the generated shape.

\section{Experiment}
\label{sec:experiment}

\subsection{Dataset}
We use the ShapeNet~\cite{chang2015shapenet} dataset to train our model and compare it with baselines.
It contains models from $55$ categories.
We split the dataset into training set ($70\%$), validation set ($10\%$), and test set ($20\%$).
We normalize each 3D model in the range of $[-1,1]^3$.
To obtain gound-truth mask silhouettes and depth maps, we render mask silhouettes and depth maps for each 3D model from $100$ random views at the resolution of $1024 \times 1024$.
The radius is fixed at $2.6$. Elevation angle and azimuth angle is sampled uniformly from $[90^{\circ}, -90^{\circ}]$ and $[0^{\circ}, 360^{\circ}]$, respectively. FOV is set to $60^{\circ}$.

\begin{figure*}[t]
\centering
\vspace{-1em}
\includegraphics[width=\linewidth]{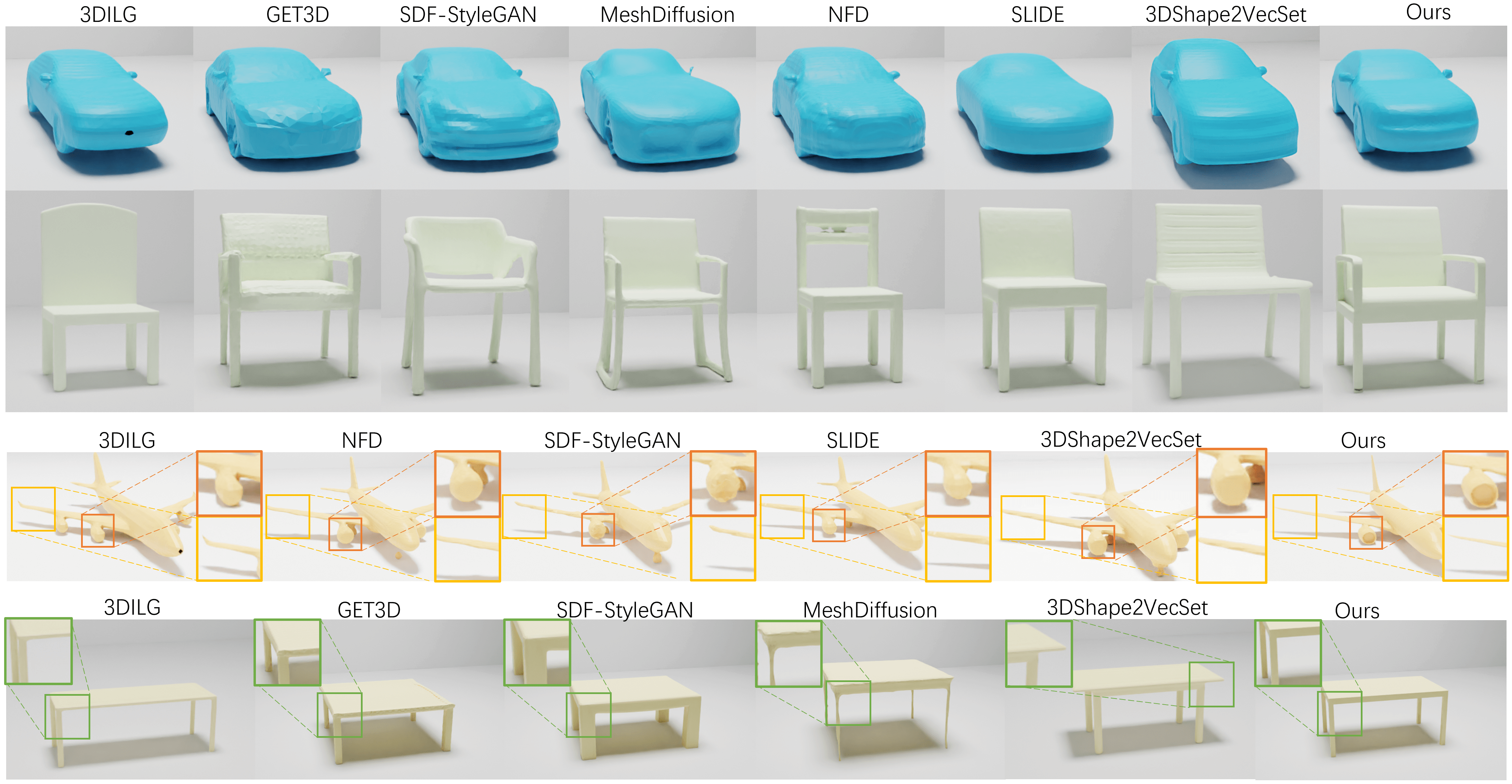}
\vspace{-1em}
\caption{Visual comparison between meshes generated by our method and baselines. Zoom in to better see the details. More qualitative results are in Appendix Section~\ref{sec: more_qualitative_results}.}
\label{fig: comparison_single_class}
\vspace{-1.5em}
\end{figure*}

\subsection{Implementation Details}

\noindent \textbf{Model Architecture.}
For the mesh autoencoder, we set the number of latent points to $N_{\text{min}}=128, N_{\text{max}}=256$.
For the point-based encoder, we use 3 layers of SA modules~\cite{lyu2023controllable}.
The size of the triplane is $3 \times 256 \times 256 \times \times 32$.
The triplane-based decoder is composed of a UNet with 3D-aware convolutions, a DMTet grid with resolution $128^3$, and a refinement module.
Detailed architecture of the mesh autoencoder is in Appendix Section~\ref{sec: appendix_mesh_autoencoder_architecture}.
For both the position DDPM and feature DDPM, we use a Transformer architecture composed of $12$ Multi-Head Self-Attettion blocks. The embedding dimension of each attention block is $512$ and the number of attention heads is $8$.
Detailed architecture of the position DDPM and feature DDPM is in Appendix Section~\ref{sec: appendix_latent_diffusion_model}.

\noindent \textbf{Training.}
We train the mesh autoencoder on all $55$ categories in ShapeNet.
We sample $N_{\text{in}}=16384$ points from the mesh surface as input point cloud.
At each training step, we randomly sample $N_{\text{view}}=8$ views out of the $100$ views in the dataset to supervise the reconstructed mesh.
The mesh autoencoder is trained in three phases for $900$ epochs with a batchsize of $128$.
Detailed training schedule and time of the mesh autoencoder are in Appendix Section~\ref{sec: appendix_mesh_autoencoder_architecture}.
For both the position DDPM and feature DDPM, we use the Adam optimizer to train them for $4000$ epochs with batchsize $128$ and learning rate $2 \times 10^{-4}$.
More training details of the position DDPM and feature DDPM are in Appendix Section~\ref{sec: appendix_latent_diffusion_model}.

\subsection{Mesh Autoencoding}
\label{sec: train_mesh_ae}
The autoencoder is trained on the training set.
We select the checkpoint with the best performance on the validation set and evaluate its performance on the test set.
We compute the IOU between rendered masks of the reconstructed meshes and ground-truth masks (averaged across the first 32 views in the dataset), and $L_2$ CD (Chamfer Distance) loss between $10^{5}$ surface points sampled from the reconstructed meshes and ground-truth meshes.
Our mesh autoencoder achieves $0.968$ IOU and $1.34 \times 10^{-3}$ CD loss.

\begin{table}[t]
\centering
\vspace{-1em}
\caption{Compare the average generation time per mesh and Shading-FID of our method and baselines. ``-'' indicates that no checkpoint is provided for that category.
To fairly compare the generation speed and quality, we use $1000$ denoising steps during inference for all DDPM-based methods, except for MeshDiffusion as it is too slow.
We use DDIM~\cite{song2020denoising} ($100$ Steps) to accelerate MeshDiffusion.
MeshDiffusion uses a $128^3$ DMTet grid (903.20s per sample) for Car and Chair, and uses a $64^3$ DMTet grid (91.35s per sample) for Airplane and Table.
}
\vspace{-1em}
\scalebox{0.85}{
\begin{tabular}{l|l|l|c|cccc}
    \toprule[1.5pt]
    \multirow{2}{*}{Method}& \multirow{2}{*}{Category} & \multirow{2}{*}{Model Type} &  Generation & \multicolumn{4}{c}{Shading-FID $\downarrow$} \\
    \cline{5-8} 
    & & & Time (s) & Airplane & Car & Chair & Table \\
    \midrule[1pt]
    SDF-StyleGAN~\cite{zheng2022sdf} & Single & GAN & 1.25 & 70.49 & 105.40 & 46.04 & 45.57\\
    GET3D~\cite{Gao2022GET3DAG} & Single & GAN & 0.12 & - & 182.07 & 66.48 & 64.06\\
    3DILG~\cite{Zhang20223DILGIL} & Multiple & Autoregressive & 9.91 & 58.18 & 152.00 & 29.71 & 52.97\\
    NFD~\cite{Shue20223DNF} & Single & DDPM & 7.93 & 54.43 & 182.58 & 42.66 & - \\
    MeshDiffusion~\cite{Liu2023MeshDiffusionSG} & Single & DDPM & 91.35, 903.20 & 134.10 & 151.76 & 76.81 & 79.59\\
    SLIDE~\cite{lyu2023controllable} & Single & DDPM & 0.20 & 85.17 & 199.84 & 43.64 & -\\
    3DShape2VecSet~\cite{zhang20233dshape2vecset} & Multiple & DDPM & 6.56 & \second{47.86} & \textbf{101.38}  & 22.92 & \second{23.90}\\
    \midrule 
    Ours & Multiple & DDPM & 1.17 & \textbf{32.10} & 121.42 & \textbf{22.16} & \textbf{16.50} \\
    \bottomrule[1.5pt]
\end{tabular}
}

\label{table: shading_fid}
\end{table}
\begin{table}[t]
\centering
\vspace{-0.5em}
\caption{1-NNA comparison between our method and baselines. 
GetMesh achieves the best 1-NNA compared even with single-category models, which demonstrates that meshes generated by GetMesh best match the distribution of the dataset.
}
\vspace{-1em}
\scalebox{0.9}{
\begin{tabular}{l|l|cccc|cccc}
    \toprule[1.5pt]
    \multirow{2}{*}{Method} & \multirow{2}{*}{Category} & \multicolumn{4}{c|}{1-NNA (CD) (Percent) $\downarrow$} & \multicolumn{4}{c}{1-NNA (EMD) (Percent) $\downarrow$}\\
    \cline{3-10}
    & & Airplane & Car & Chair & Table & Airplane & Car & Chair & Table\\
    \midrule[1pt]
    SDF-StyleGAN~\cite{zheng2022sdf} & Single & 88.10 & 95.20 & 65.22 & 75.18 &91.75 & 94.65 & 70.27 & 75.08\\
    GET3D~\cite{Gao2022GET3DAG} &Single & - & 97.00 & 68.97 & 68.37 & - & 92.90 & 65.16 & 67.76\\
    3DILG~\cite{Zhang20223DILGIL} &Multiple & 85.20 & 95.25 & 74.27 & 82.43 & 88.15 & 94.30 & 73.62 & 81.43\\
    NFD~\cite{Shue20223DNF} &Single & 70.75 & 86.8 & 53.5 & - & 77.3 & 87.85 & 55.06 & - \\
    MeshDiffusion~\cite{Liu2023MeshDiffusionSG} &Single & 73.60 & 89.20 & 67.77 & 57.81 &74.60 & 88.10 & 67.86 & 60.21\\
    SLIDE~\cite{lyu2023controllable} &Single &80.20 & 91.95 & 59.41 & - &80.25 & 92.55 & 61.01 & -\\
    3DShape2VecSet~\cite{zhang20233dshape2vecset} &Multiple & 72.4 & 89.45 & 60.66 & 56.61 & 78.25 & 89.4 & 60.91 & 58.66 \\
    \midrule
    Ours & Multiple & \textbf{69.95} & \textbf{84.95} & \textbf{52.05} & \textbf{52.8} & \textbf{69.20} & \textbf{81.65} & \textbf{53.6} & \textbf{52.1}\\
    \bottomrule[1.5pt]
\end{tabular}
}

\label{table: 1_nna_baseline_and_ours}
\vspace{-1.5em}
\end{table}
\subsection{Mesh Generation}
\begin{table}[t]
\centering
\vspace{-1em}
\caption{MMD comparison between our method and baselines. 
GetMesh achieves good MMD compared even with single-category models, which demonstrates the high generation quality of GetMesh.
}
\vspace{-1em}
\scalebox{0.9}{
\begin{tabular}{l|l|cccc|cccc}
    \toprule[1.5pt]
    \multirow{2}{*}{Method} & \multirow{2}{*}{Category} & \multicolumn{4}{c|}{MMD (CD) $\times 1000$ $\downarrow$} & \multicolumn{4}{c}{MMD (EMD) $\times 100$ $\downarrow$}\\
    \cline{3-10}
    & & Airplane & Car & Chair & Table & Airplane & Car & Chair & Table\\
    \midrule[1pt]
    SDF-StyleGAN~\cite{zheng2022sdf} & Single &4.31 & 4.87 & 16.42 & 23.05 & 11.44 & 8.82 & 16.95 & 17.17 \\
    GET3D~\cite{Gao2022GET3DAG} & Single &- & 4.97 & 17.16 & 21.08 & - & 8.67 & 16.61 & 16.57 \\
    3DILG~\cite{Zhang20223DILGIL} &Multiple & 4.49 & 5.18 & 18.56 & 30.13 & 9.33 & 8.80 & 17.34 & 19.39 \\
    NFD~\cite{Shue20223DNF} & Single &3.08 & 4.17 & \textbf{14.44} & - & 8.38 & 8.15 & \textbf{15.00} & - \\
    MeshDiffusion~\cite{Liu2023MeshDiffusionSG} &Single & \textbf{3.02} & 4.79 & 17.11 & 18.68 & \textbf{8.10} & 8.88 & 17.17 & 15.82 \\
    SLIDE~\cite{lyu2023controllable} &Single & 3.60 & 4.43 & 15.03 & - & 8.64 &    8.38 & 15.88 & - \\
    3DShape2VecSet~\cite{zhang20233dshape2vecset} &Multiple & 3.16 & 4.20 & 15.51 & 18.69 & 8.79 & 8.41& 16.20 & 15.88 \\
    \midrule
    Ours &Multiple & 3.38 & \textbf{4.01} & 14.55 & \textbf{17.86} & 8.31 & \textbf{8.00} & 15.72 & \textbf{15.34} \\
    \bottomrule[1.5pt]
\end{tabular}
}
\label{table: mmd_baseline_and_ours}
\end{table}
\begin{table}[t]
\centering
\vspace{-0.5em}
\caption{Coverage comparison between our method and baselines. 
GetMesh achieves high Coverage compared even with single-category models, which demonstrates the high generation diversity of GetMesh.}
\vspace{-1em}
\scalebox{0.9}{
\begin{tabular}{l|l|cccc|cccc}
    \toprule[1.5pt]
    \multirow{2}{*}{Method} & \multirow{2}{*}{Category} & \multicolumn{4}{c|}{Coverage (CD) (Percent) $\uparrow$} & \multicolumn{4}{c}{Coverage (EMD) (Percent) $\uparrow$}\\
    \cline{3-10}
    & & Airplane & Car & Chair & Table & Airplane & Car & Chair & Table\\
    \midrule[1pt]
    SDF-StyleGAN~\cite{zheng2022sdf} &Single & 32.5 & 20.0 & 43.24 & 38.13 & 21.7& 19.8 & 38.43 & 35.83\\
    GET3D~\cite{Gao2022GET3DAG} &Single & - & 20.5 & 45.95 & 44.04 & - & 23.4 & 47.75 & 45.95 \\
    3DILG~\cite{Zhang20223DILGIL} &Multiple & 35.5 & 13.6 & 36.54 & 24.22 & 28.3 & 13.4 & 37.84 & 27.23 \\
    NFD~\cite{Shue20223DNF} &Single & 46.4 & \textbf{29.5} & 46.75 & - & 36.8 & 27.0 & 46.65 & - \\
    MeshDiffusion~\cite{Liu2023MeshDiffusionSG} &Single & \textbf{48.0} & 25.2 & 40.14 & 49.05 & 43.8 & 28.3 & 42.54 & 45.25 \\
    SLIDE~\cite{lyu2023controllable} &Single &40.8 & 26.3 & 46.75 & - & 38.1 & 23.7 & 46.65 & - \\
    3DShape2VecSet~\cite{zhang20233dshape2vecset} &Multiple & 47.0 & 28.4 & \textbf{50.45} & 49.95 & 43.1 & 27.6 & 49.35 & 50.25 \\
    \midrule
    Ours &Multiple & 46.1 & 29.0 & 49.35 & \textbf{51.45} & \textbf{50.5} & \textbf{30.4} & \textbf{50.35} & \textbf{51.15 } \\
    \bottomrule[1.5pt]
\end{tabular}
}
\label{table: cov_baseline_and_ours}
\vspace{-1.5em}
\end{table}

We train the position DDPM and the feature DDPM in the latent space of the pre-trained autoencoder in the previous section.
We train class-conditional DDPMs on the $55$ categories.
Each class is associated with a learnable token and the token is appended to the sequence of positions and features of the latent points during training.
We randomly drop the class label to null with $20\%$ probability during training to enable classifier-free~\cite{ho2022classifier} guidance in the sampling phase, and novel shape generation and editing beyond the $55$ categories in ShapeNet.

We compare our method with both multi-category generative models~\cite{Zhang20223DILGIL, zhang20233dshape2vecset} and single-category generative models~\cite{Gao2022GET3DAG, zheng2022sdf, Shue20223DNF, Liu2023MeshDiffusionSG, lyu2023controllable}.
We use Shading FID~\cite{zheng2022sdf}, 1-NNA~\cite{Yang2019PointFlow3P}, Minimum Matching Distance (MMD~\cite{achlioptas2018learning}) and Coverage~\cite{achlioptas2018learning} 
to evaluate our method and baselines.
More evaluation details are in Appendix Section~\ref{sec: evaluation_details}.
Results are shown in Table~\ref{table: shading_fid}, Table~\ref{table: 1_nna_baseline_and_ours}, Table~\ref{table: mmd_baseline_and_ours} and Table~\ref{table: cov_baseline_and_ours}, respectively.
We can see that our model achieves the best 1-NNA in all cases and the best Shading-FID in most cases even compared with single-category models.
Both metrics measure the distance between the distribution of the generated meshes and meshes in the reference dataset.
Therefore, the experiments demonstrate that GetMesh better learns the distribution of meshes in the dataset compared with baselines.
MMD and Coverage explicitly measure the quality and diversity of generated meshes, respectively.
We can see that GetMesh achieves highly competitive MMD and Coverage even compared with single-category models, which demonstrates the superior generation quality and diversity of GetMesh.

Besides quantitative comparisons, we also qualitatively compare meshes generated by GetMesh and baselines in Figure~\ref{fig: comparison_single_class}.
We can see that our method generates meshes with sharper edges and smoother surfaces.
Our method also generates thin structures such as airplane wings and tails quite well.

\begin{figure}[t]
\centering
\vspace{-1em}
\includegraphics[width=0.95\linewidth]{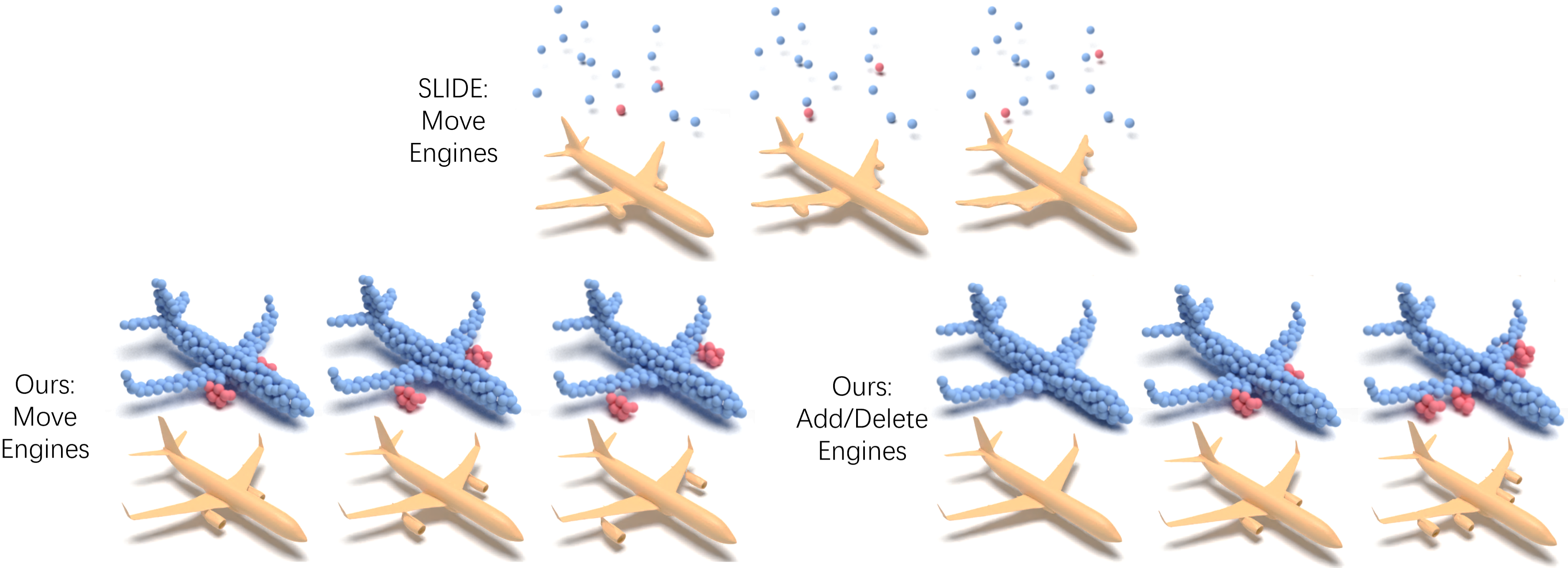}
\caption{Compare our method with SLIDE in terms of controllable generation. Our method can perform more delicate control on the generated shape and enables adding or deleting shape parts.}
\label{fig: compare_with_slide}
\end{figure}

\noindent \textbf{Generation speed.}
We compare the average generation time per sample of our method and baselines tested on a single NVIDIA A100 GPU in Table~\ref{table: shading_fid}.
We can see that GetMesh is much faster than other DDPM-based method.
That is because we train diffusion models on a much more compact point-based representation, which contains $128\sim 256$ points.
Note that triplanes are only used in our autoencoder to reconstruct high-quality meshes, not used to train diffusion models.
On the other hand, MeshDiffusion trains diffusion models on the DMTet grid, which contains $277410$ vertices for a $128^3$ resolution grid.
NFD trains diffusion models on triplanes of resolution $128$, which contains $128\times 128 \times 3 = 49152$ pixels.
3DShape2VecSet trains diffusion models on a set of $512$ vectors.
This explains why GetMesh is much faster than these DDPM-based methods.
Although SLIDE~\cite{lyu2023controllable} is faster than our method by using fewer points ($16$ points), GetMesh enables much more accurate control over the generated shapes than SLIDE by using relatively more points as shown in Figure~\ref{fig: compare_with_slide}.
In addition, GetMesh generates much higher quality meshes than SLIDE by using a triplane-based mesh decoder compared to SLIDE's Poisson-based surface reconstruction method~\cite{Peng2021ShapeAP}.
The generation speed of GetMesh is also competitive compared with other non-DDPM-based methods.

\begin{figure}[t]
\centering
\vspace{-1em}
\includegraphics[width=1\linewidth]{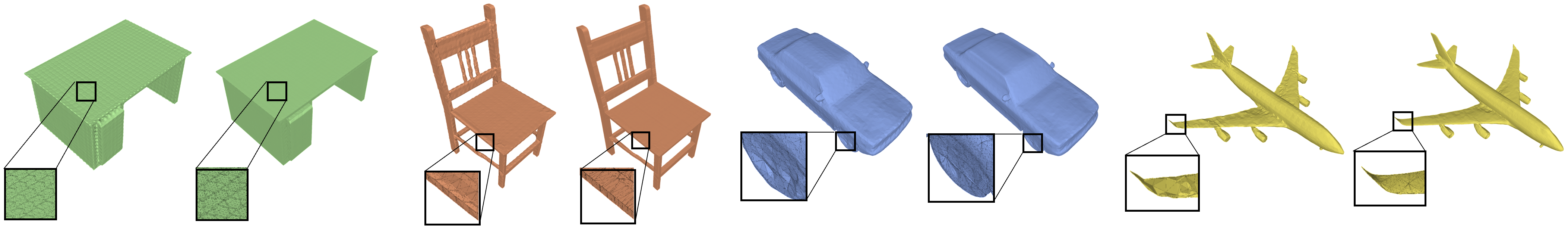}
\caption{Compare meshes reconstructed by autoencoders with and without the refinement module. For each pair of meshes, the left one is without the refinement module, and the right one is with the module. Zoom in to see more details.}
\label{fig: compare_coarse_and_refined}
\vspace{-0.5em}
\end{figure}

\begin{table}[tb]
\centering
\caption{Ablation study of the number of latent points in the latent space. Shading FID is reported. }
\vspace{-0.5em}
\scalebox{0.85}{
\begin{tabular}{cc|cccc}
    \toprule[1.5pt]
    $N_{\text{min}}$ & $N_{\text{max}}$ & Airplane & Car & Chair & Table\\
    \midrule[1pt]
    32 & 64 & 206.57 & 261.49 & 224.71 & 228.43 \\
    128 & 256 & \textbf{32.10} & 121.42 & \textbf{24.59} & \textbf{16.50} \\
    512 & 1024 & 44.32 & \textbf{113.22} & 63.25 & 36.03 \\
    2048 & 4096 & 237.36 & 163.16 & 162.83 & 117.55 \\
    \bottomrule[1.5pt]
\end{tabular}
}
\label{table: ablation_shading_fid}
\vspace{-1.5em}
\end{table}
\subsection{Ablation Study}

We conduct an ablation study on the number of latent points in the latent space.
The performance of the mesh autoencoders with different numbers of latent points is shown in Appendix Section~\ref{sec: ablation_study}.
The mesh autoencoder with $N \in [512, 1024]$ achieves the best reconstruction performance.
Next, we train latent diffusion models in the latent space of these autoencoders.
We report the generation performance of the latent diffusion models in Table~\ref{table: ablation_shading_fid} and more results are in Appendix Section~\ref{sec: ablation_study}.
We can see that the diffusion model with $N \in [128, 256]$ achieves the best generation performance in most cases, and the diffusion model with $N \in [512, 1024]$ achieves relatively good generation performance as well.
On the other hand, the performance of the diffusion models with too many ($N \in [2048, 4096]$) or too few ($N \in [32, 64]$) latent points is significantly worse than the other two diffusion models.
We need a moderate number of latent points ($128 \sim 1024$) to train latent diffusion models with good generation performance.

Next, we ablate our refinement module in the mesh autoencoder.
Without the refinement module, the reconstruction performance of the mesh autoencoder ($N \in [128, 256]$) drops from $0.968$ IOU and $1.34 \times 10^{-3}$ CD loss to $0.961$ IOU and $1.46 \times 10^{-3}$ CD loss.
Figure~\ref{fig: compare_coarse_and_refined} also provides a qualitative comparison between meshes reconstructed with and without the refinement module.
We can see that the refinement module greatly mitigates the artifacts in the coarse meshes extracted from DMTet.
It makes the mesh surface smoother and the edges sharper.

\begin{figure*}[t]
\centering
\vspace{-1em}
\includegraphics[width=1\linewidth]{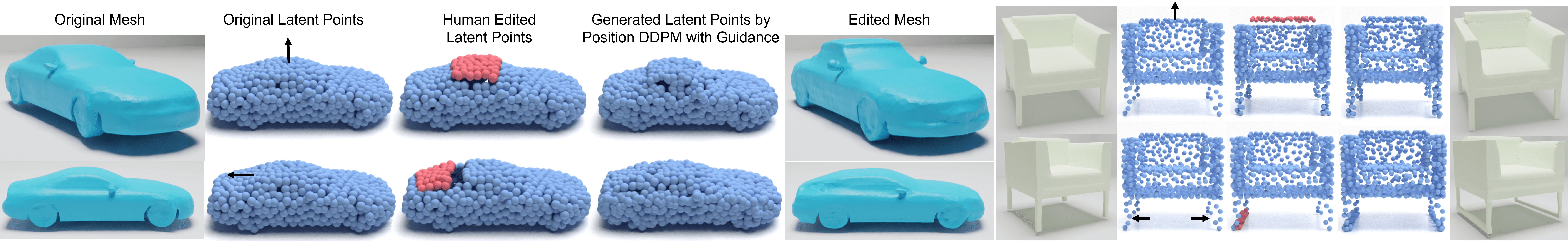}
\caption{Human-edited latent points could have holes, missing parts, or other flaws. We develop a guidance method for the position DDPM to generate latent points that are consistent with the edited latent points, but mitigate flaws in the edited latent points. This guidance method makes mesh manipulation through latent point positions more robust and convenient.}
\label{fig: position_ddpm_guided_sampling}
\vspace{-1.5em}
\end{figure*}

\subsection{Controllable Mesh Generation}
\label{sec: controllable_mesh_generation}

Our method enables highly controllable 3D shape generation and flexible mesh manipulation by adjusting the number, positions, or features of the latent points.
We demonstrate the controllable generation ability of the model with $N \in [512, 1024]$ in this section, and more examples of the model with $N \in [128, 256]$ are shown in Appendix Section~\ref{sec: controllable_shape_generation}.

\begin{figure*}[t]
\centering
\vspace{-1em}
\includegraphics[width=1\linewidth]{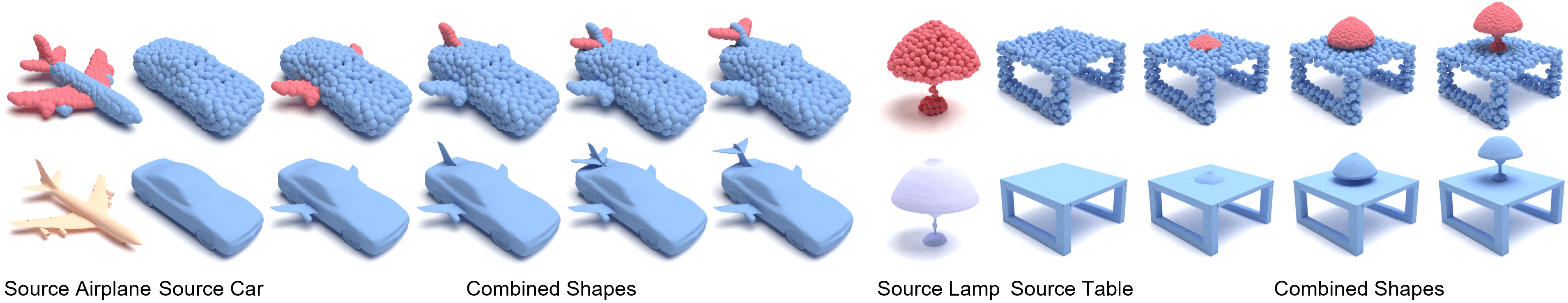}
\caption{We can combine different shapes to form novel shapes by combining their latent points. Our models can properly handle the joints of different parts and output watertight meshes.}
\label{fig: shape_combination}
\vspace{-1em}
\end{figure*}

\begin{figure}[t]
\begin{minipage}{0.45\textwidth}
\centering
\includegraphics[width=1\linewidth]{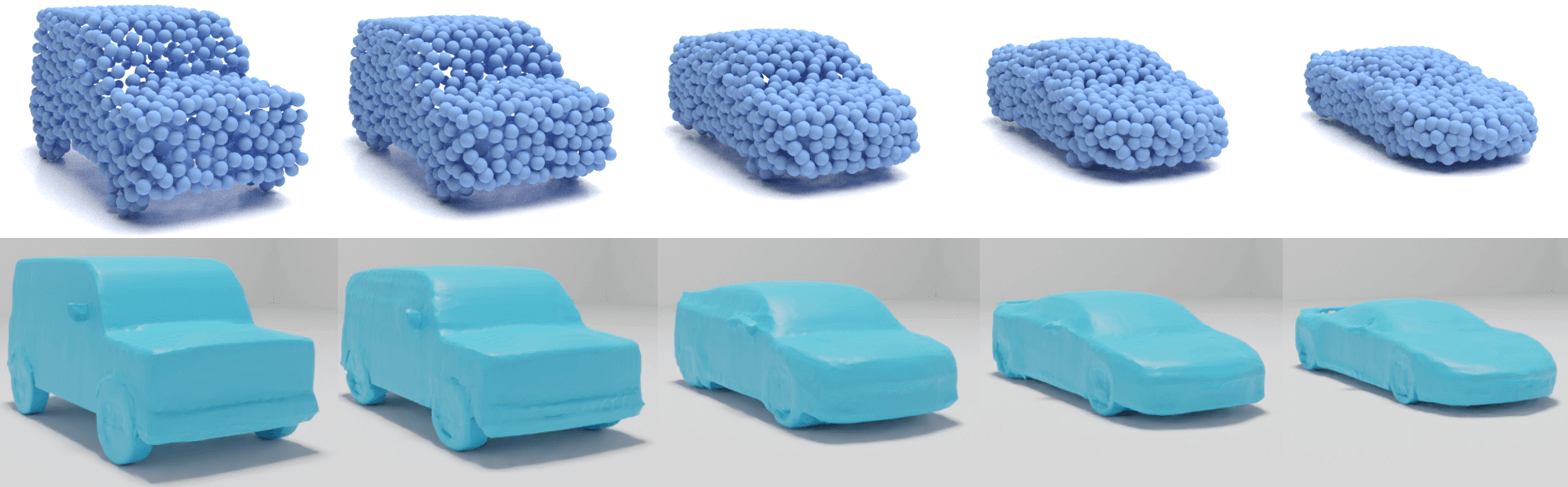}
\caption{Shape interpolation.
}
\label{fig: shape_interpolation}
\end{minipage}
\hspace{0.01\textwidth}
\begin{minipage}{0.525\textwidth}
\centering
\includegraphics[width=1\linewidth]{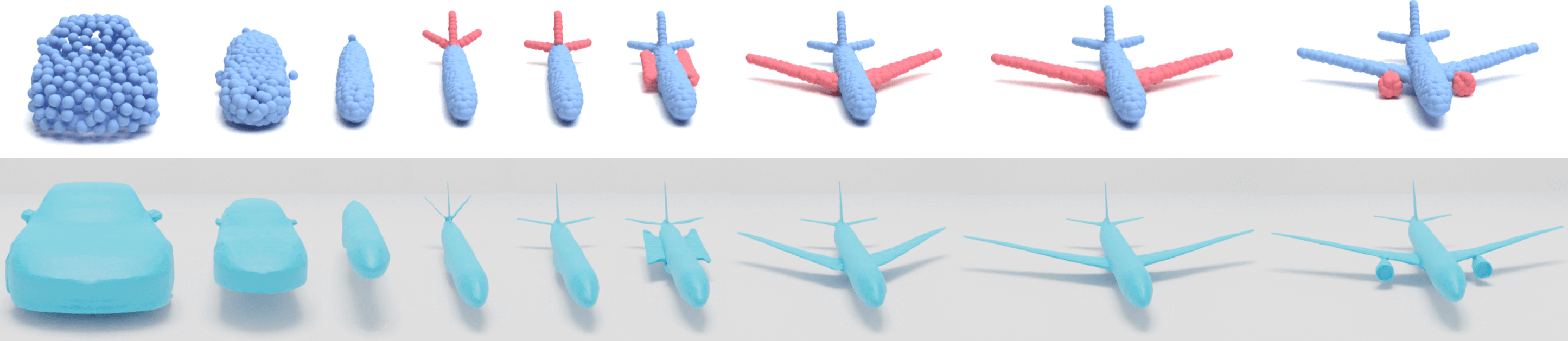}
\caption{Turning a car into an airplane.}
\label{fig: animation_car_to_airplane}
\end{minipage}
\vspace{-1.5em}
\end{figure}

\begin{figure*}[t]
\centering
\vspace{-1em}
\includegraphics[width=1\linewidth]{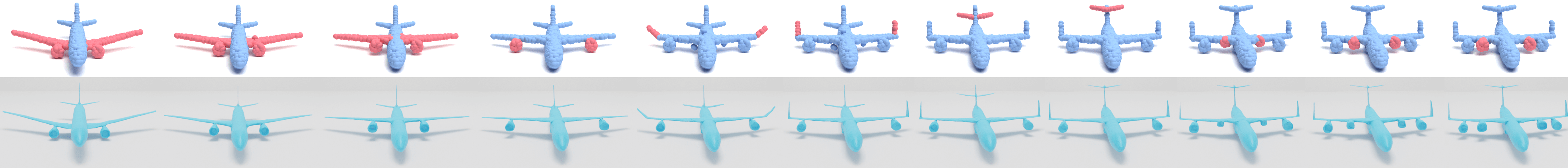}
\caption{Turning a twin-engine passenger airplane into a four-engine transport airplane.}
\label{fig: animation_passenger_to_transport_airplane}
\vspace{-0.5em}
\end{figure*}

\noindent \textbf{Controllable mesh generation.}
We can use the positions of the latent points to control the shape of generated meshes.
We compare our method with SLIDE~\cite{lyu2023controllable}, which also supports latent point-based shape manipulation.
Our models have more latent points than SLIDE ($16$ points), therefore, we can perform more delicate controls over the generated mesh.
As shown in Figure~\ref{fig: compare_with_slide}, we can control the positions of the engines of the generated airplane while SLIDE struggles to control.
In addition, our models support varying numbers of latent points, and thus we can delete or add a part of a shape.
Figure~\ref{fig: compare_with_slide} shows that we can delete or add engines to the airplane, which SLIDE can not achieve since it uses a fixed number of latent points.

We also develop a guided-sampling method for the position DDPM to facilitate controllable mesh generation and manipulation.
Human-edited latent points may have flaws in many cases. 
Therefore, we need to develop a method that reflects the editing effect of a user's intention, while mitigating the pitfalls of human editing.
The main idea of our method is to leverage the position DDPM's ability to generate an arbitrary number of latent points and use the edited latent points to guide the sampling process of the position DDPM.
The guidance method is inspired by the one proposed in~\cite{fei2023generative} and is explained in Appendix Section~\ref{sec: position_ddpm_guided_sampling}.
In Figure~\ref{fig: position_ddpm_guided_sampling}, we demonstrate that our guidance method can make the position DDPM generate latent points that are consistent with the edited latent points but mitigate their flaws, and thus facilitate the process of shape manipulation.

\noindent \textbf{Shape combination.}
Since our models support varying numbers of latent points, we can directly combine different shapes together to form novel shapes.
We can combine the latent points and regenerate features using the feature DDPM or simply remaining the original features, and then use the mesh autoencoder to decode the combined latent points to a mesh.
Figure~\ref{fig: shape_combination} shows that our models can properly handle the joints of different parts.

\noindent \textbf{Shape interpolation.}
Shape interpolation is straightforward using latent points and their features.
For two shapes represented by latent points $\vx_1,  \vx_2 \in \mathbb{R}^{N \times 3}$ and latent features $\vy_1,  \vy_2 \in \mathbb{R}^{N \times D}$, 
we find the optimal bijection $\phi^{*} : \vx_1 \longrightarrow \vx_2$.
\vspace{-0.5em}
\begin{align}
    \phi^{*} = \argmin_{\phi} \Sigma_{i=1}^{N} ||\vx_1^i - \phi(\vx_1^i)||_2,
\vspace{-0.5em}
\end{align}
where $\vx_1^i$ is the $i$-th point in $\vx_1$.
Based on this bijection $\phi^{*}$, we can interpolate the corresponding points and their features between $\vx_1$ and $\vx_2$.
Figure~\ref{fig: shape_interpolation} gives an example of shape interpolation.

\noindent \textbf{Shape animation.}
Traditional mesh animation methods drive 3D meshes by deforming the mesh.
The vertices of the mesh are deformed while the topology of the mesh is fixed, namely, the connections of the vertices are fixed.
This could prevent them from performing some complex shape animations that require topology changes.
On the other hand, our method can generate meshes of arbitrary topology given the latent points.
Therefore, we can use the latent points to drive the mesh.
Specifically, we can use a sequence of latent point positions to create an animation involving complex shape transformations.
Figure~\ref{fig: animation_passenger_to_transport_airplane} and Figure~\ref{fig: animation_car_to_airplane} are two examples.
The complete video can be found in the supplementary material.
We think that our method provides a complementary approach to existing mesh animation methods and enables more creative and complex animations of 3D meshes.

\begin{figure}[t]
\centering
\includegraphics[width=0.6\linewidth]{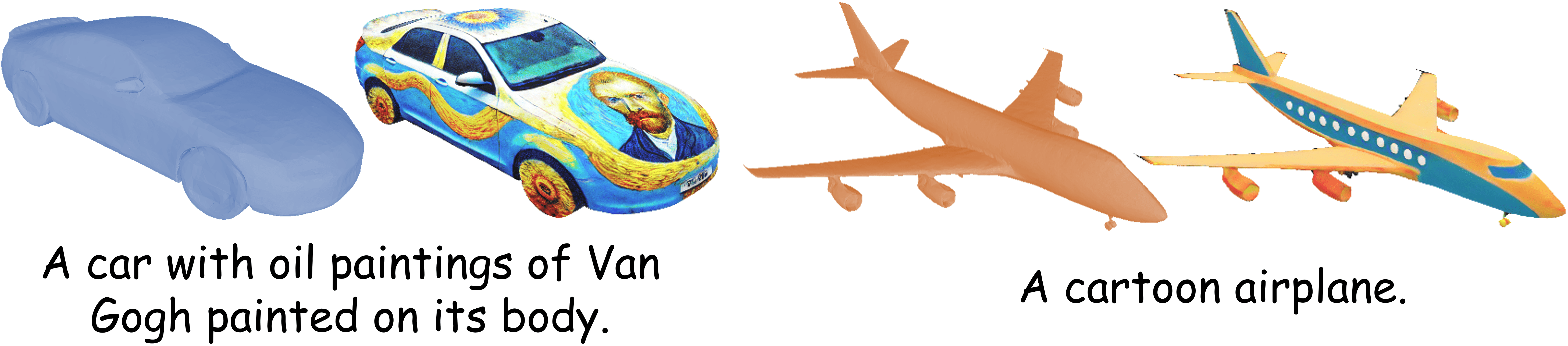}
\caption{We use MATLABER~\cite{xu2023matlaber} to generate materials for meshes generated by our method. More examples are in Appendix Section~\ref{sec: more_qualitative_results}. 
Note that texture or material generation is NOT the focus or contribution of this work. 
We merely intend to show that our method can be seamlessly combined with off-the-shelf methods to obtain meshes with textures or materials.}
\label{fig: textures_with_novel_prompts}
\vspace{-1.5em}
\end{figure}

\noindent \textbf{Textured mesh generation.}
It is possible to paint our generated meshes with some off-the-shelf methods.
We use MATLABER~\cite{xu2023matlaber} to colorize our generated meshes, which can generate materials for meshes by leveraging a powerful text-to-image diffusion model.
Some examples are shown in Figure~\ref{fig: textures_with_novel_prompts}, and more examples are in Appendix Section~\ref{sec: more_qualitative_results}.

\vspace{-0.5em}
\section{Limitations}
\label{sec:limitation}
\vspace{-0.5em}

While GetMesh has made significant progress in 3D generation quality and controllability, it still has some limitations.
Firstly, training GetMesh requires ground-truth 3D data, which is quite expensive to acquire compared with 2D images.
In light of recent works~\cite{Poole2022DreamFusionTU, Lin2022Magic3DHT, Xu2022Dream3DZT, Chen2023Fantasia3DDG, Wang2023ProlificDreamerHA} that generate 3D assets using 2D Text-to-Image diffusion models trained on large-scale image datasets,
it is possible to combine the 3D priors in GetMesh with 2D priors in Text-to-Image diffusion models to achieve more diverse and high-quality 3D generation.
In addition, GetMesh is only validated on the ShapeNet~\cite{chang2015shapenet} dataset due to limited computational resources.
We plan to further verify the scalability of GetMesh on larger-scale datasets such as Objaverse~\cite{deitke2023objaverse} in the future.

\vspace{-0.5em}
\section{Conclusion}
\label{sec:conclusion}
\vspace{-0.5em}

In this paper, we propose \name, a multi-category generative model that enables both high-quality mesh generation and flexible control over the generated shape.
It combines the advantages of both point-based representation and triplane-based representation.
The triplane-based representation associated with a decoder and a refinement module enables us to reconstruct high-quality meshes from the latent representation.
By adjusting the number, positions or features of the latent points, we can intuitively and robustly change global/local topologies of meshes, add/remove mesh parts, as well as combine mesh parts across different instances/categories.
\bibliographystyle{splncs04}
\bibliography{main}

\begin{thebibliography}{10}
\providecommand{\url}[1]{\texttt{#1}}
\providecommand{\urlprefix}{URL }
\providecommand{\doi}[1]{https://doi.org/#1}

\bibitem{achlioptas2018learning}
Achlioptas, P., Diamanti, O., Mitliagkas, I., Guibas, L.: Learning representations and generative models for 3d point clouds. In: International conference on machine learning. pp. 40--49. PMLR (2018)

\bibitem{chang2015shapenet}
Chang, A.X., Funkhouser, T., Guibas, L., Hanrahan, P., Huang, Q., Li, Z., Savarese, S., Savva, M., Song, S., Su, H., et~al.: Shapenet: An information-rich 3d model repository. arXiv preprint arXiv:1512.03012  (2015)

\bibitem{Chen2023Fantasia3DDG}
Chen, R., Chen, Y., Jiao, N., Jia, K.: Fantasia3d: Disentangling geometry and appearance for high-quality text-to-3d content creation. ArXiv  \textbf{abs/2303.13873} (2023), \url{https://api.semanticscholar.org/CorpusID:257757213}

\bibitem{Chou2022DiffusionSDFCG}
Chou, G., Bahat, Y., Heide, F.: Diffusionsdf: Conditional generative modeling of signed distance functions. ArXiv  \textbf{abs/2211.13757} (2022), \url{https://api.semanticscholar.org/CorpusID:254017862}

\bibitem{deitke2023objaverse}
Deitke, M., Schwenk, D., Salvador, J., Weihs, L., Michel, O., VanderBilt, E., Schmidt, L., Ehsani, K., Kembhavi, A., Farhadi, A.: Objaverse: A universe of annotated 3d objects. In: Proceedings of the IEEE/CVF Conference on Computer Vision and Pattern Recognition. pp. 13142--13153 (2023)

\bibitem{Dhariwal2021DiffusionMB}
Dhariwal, P., Nichol, A.: Diffusion models beat gans on image synthesis. ArXiv  \textbf{abs/2105.05233} (2021), \url{https://api.semanticscholar.org/CorpusID:234357997}

\bibitem{fei2023generative}
Fei, B., Lyu, Z., Pan, L., Zhang, J., Yang, W., Luo, T., Zhang, B., Dai, B.: Generative diffusion prior for unified image restoration and enhancement. In: Proceedings of the IEEE/CVF Conference on Computer Vision and Pattern Recognition. pp. 9935--9946 (2023)

\bibitem{Gao2022GET3DAG}
Gao, J., Shen, T., Wang, Z., Chen, W., Yin, K., Li, D., Litany, O., Gojcic, Z., Fidler, S.: Get3d: A generative model of high quality 3d textured shapes learned from images. ArXiv  \textbf{abs/2209.11163} (2022), \url{https://api.semanticscholar.org/CorpusID:252438648}

\bibitem{gupta20233dgen}
Gupta, A., Xiong, W., Nie, Y., Jones, I., O{\u{g}}uz, B.: 3dgen: Triplane latent diffusion for textured mesh generation. arXiv preprint arXiv:2303.05371  (2023)

\bibitem{ho2020denoising}
Ho, J., Jain, A., Abbeel, P.: Denoising diffusion probabilistic models. Advances in neural information processing systems  \textbf{33},  6840--6851 (2020)

\bibitem{Ho2021CascadedDM}
Ho, J., Saharia, C., Chan, W., Fleet, D.J., Norouzi, M., Salimans, T.: Cascaded diffusion models for high fidelity image generation. J. Mach. Learn. Res.  \textbf{23},  47:1--47:33 (2021), \url{https://api.semanticscholar.org/CorpusID:235619773}

\bibitem{ho2022classifier}
Ho, J., Salimans, T.: Classifier-free diffusion guidance. arXiv preprint arXiv:2207.12598  (2022)

\bibitem{Jeong2021DiffTTSAD}
Jeong, M., Kim, H., Cheon, S.J., Choi, B.J., Kim, N.S.: Diff-tts: A denoising diffusion model for text-to-speech. In: Interspeech (2021), \url{https://api.semanticscholar.org/CorpusID:233025015}

\bibitem{Kong2020DiffWaveAV}
Kong, Z., Ping, W., Huang, J., Zhao, K., Catanzaro, B.: Diffwave: A versatile diffusion model for audio synthesis. ArXiv  \textbf{abs/2009.09761} (2020), \url{https://api.semanticscholar.org/CorpusID:221818900}

\bibitem{Li2022DiffusionSDFTV}
Li, M., Duan, Y., Zhou, J., Lu, J.: Diffusion-sdf: Text-to-shape via voxelized diffusion. 2023 IEEE/CVF Conference on Computer Vision and Pattern Recognition (CVPR) pp. 12642--12651 (2022), \url{https://api.semanticscholar.org/CorpusID:254366593}

\bibitem{Lin2022Magic3DHT}
Lin, C.H., Gao, J., Tang, L., Takikawa, T., Zeng, X., Huang, X., Kreis, K., Fidler, S., Liu, M.Y., Lin, T.Y.: Magic3d: High-resolution text-to-3d content creation. 2023 IEEE/CVF Conference on Computer Vision and Pattern Recognition (CVPR) pp. 300--309 (2022), \url{https://api.semanticscholar.org/CorpusID:253708074}

\bibitem{Liu2023MeshDiffusionSG}
Liu, Z., Feng, Y., Black, M.J., Nowrouzezahrai, D., Paull, L., yu~Liu, W.: Meshdiffusion: Score-based generative 3d mesh modeling. ArXiv  \textbf{abs/2303.08133} (2023), \url{https://api.semanticscholar.org/CorpusID:257505014}

\bibitem{Lorensen1987MarchingCA}
Lorensen, W.E., Cline, H.E.: Marching cubes: A high resolution 3d surface construction algorithm. Proceedings of the 14th annual conference on Computer graphics and interactive techniques  (1987), \url{https://api.semanticscholar.org/CorpusID:15545924}

\bibitem{Luo2021DiffusionPM}
Luo, S., Hu, W.: Diffusion probabilistic models for 3d point cloud generation. 2021 IEEE/CVF Conference on Computer Vision and Pattern Recognition (CVPR) pp. 2836--2844 (2021), \url{https://api.semanticscholar.org/CorpusID:232092778}

\bibitem{lyu2021conditional}
Lyu, Z., Kong, Z., Xu, X., Pan, L., Lin, D.: A conditional point diffusion-refinement paradigm for 3d point cloud completion. arXiv preprint arXiv:2112.03530  (2021)

\bibitem{lyu2023controllable}
Lyu, Z., Wang, J., An, Y., Zhang, Y., Lin, D., Dai, B.: Controllable mesh generation through sparse latent point diffusion models. In: Proceedings of the IEEE/CVF Conference on Computer Vision and Pattern Recognition. pp. 271--280 (2023)

\bibitem{Mescheder2018OccupancyNL}
Mescheder, L.M., Oechsle, M., Niemeyer, M., Nowozin, S., Geiger, A.: Occupancy networks: Learning 3d reconstruction in function space. 2019 IEEE/CVF Conference on Computer Vision and Pattern Recognition (CVPR) pp. 4455--4465 (2018), \url{https://api.semanticscholar.org/CorpusID:54465161}

\bibitem{ben2021nerf}
Mildenhall, B., Srinivasan, P.P., Tancik, M., Barron, J.T., Ramamoorthi, R., Ng, R.: Nerf: Representing scenes as neural radiance fields for view synthesis. Commun. ACM  \textbf{65}(1),  99–106 (dec 2021). \doi{10.1145/3503250}, \url{https://doi.org/10.1145/3503250}

\bibitem{Nam20223DLDMNI}
Nam, G., Khlifi, M., Rodriguez, A., Tono, A., Zhou, L., Guerrero, P.: 3d-ldm: Neural implicit 3d shape generation with latent diffusion models. ArXiv  \textbf{abs/2212.00842} (2022), \url{https://api.semanticscholar.org/CorpusID:254220714}

\bibitem{Nichol2021ImprovedDD}
Nichol, A., Dhariwal, P.: Improved denoising diffusion probabilistic models. ArXiv  \textbf{abs/2102.09672} (2021), \url{https://api.semanticscholar.org/CorpusID:231979499}

\bibitem{Nichol2022PointEAS}
Nichol, A., Jun, H., Dhariwal, P., Mishkin, P., Chen, M.: Point-e: A system for generating 3d point clouds from complex prompts. ArXiv  \textbf{abs/2212.08751} (2022), \url{https://api.semanticscholar.org/CorpusID:254854214}

\bibitem{pang2022masked}
Pang, Y., Wang, W., Tay, F.E., Liu, W., Tian, Y., Yuan, L.: Masked autoencoders for point cloud self-supervised learning. In: European conference on computer vision. pp. 604--621. Springer (2022)

\bibitem{Park2019DeepSDFLC}
Park, J.J., Florence, P.R., Straub, J., Newcombe, R.A., Lovegrove, S.: Deepsdf: Learning continuous signed distance functions for shape representation. 2019 IEEE/CVF Conference on Computer Vision and Pattern Recognition (CVPR) pp. 165--174 (2019), \url{https://api.semanticscholar.org/CorpusID:58007025}

\bibitem{Peng2021ShapeAP}
Peng, S., Jiang, C.M., Liao, Y., Niemeyer, M., Pollefeys, M., Geiger, A.: Shape as points: A differentiable poisson solver. In: Neural Information Processing Systems (2021), \url{https://api.semanticscholar.org/CorpusID:235358422}

\bibitem{Poole2022DreamFusionTU}
Poole, B., Jain, A., Barron, J.T., Mildenhall, B.: Dreamfusion: Text-to-3d using 2d diffusion. ArXiv  \textbf{abs/2209.14988} (2022), \url{https://api.semanticscholar.org/CorpusID:252596091}

\bibitem{Popov2021GradTTSAD}
Popov, V., Vovk, I., Gogoryan, V., Sadekova, T., Kudinov, M.A.: Grad-tts: A diffusion probabilistic model for text-to-speech. In: International Conference on Machine Learning (2021), \url{https://api.semanticscholar.org/CorpusID:234483016}

\bibitem{qi2017pointnet++}
Qi, C.R., Yi, L., Su, H., Guibas, L.J.: Pointnet++: Deep hierarchical feature learning on point sets in a metric space. Advances in neural information processing systems  \textbf{30} (2017)

\bibitem{rombach2022high}
Rombach, R., Blattmann, A., Lorenz, D., Esser, P., Ommer, B.: High-resolution image synthesis with latent diffusion models. In: Proceedings of the IEEE/CVF Conference on Computer Vision and Pattern Recognition. pp. 10684--10695 (2022)

\bibitem{shen2021deep}
Shen, T., Gao, J., Yin, K., Liu, M.Y., Fidler, S.: Deep marching tetrahedra: a hybrid representation for high-resolution 3d shape synthesis. Advances in Neural Information Processing Systems  \textbf{34},  6087--6101 (2021)

\bibitem{Shue20223DNF}
Shue, J., Chan, E., Po, R., Ankner, Z., Wu, J., Wetzstein, G.: 3d neural field generation using triplane diffusion. 2023 IEEE/CVF Conference on Computer Vision and Pattern Recognition (CVPR) pp. 20875--20886 (2022), \url{https://api.semanticscholar.org/CorpusID:254095843}

\bibitem{song2020denoising}
Song, J., Meng, C., Ermon, S.: Denoising diffusion implicit models. arXiv preprint arXiv:2010.02502  (2020)

\bibitem{vahdat2021score}
Vahdat, A., Kreis, K., Kautz, J.: Score-based generative modeling in latent space. Advances in Neural Information Processing Systems  \textbf{34},  11287--11302 (2021)

\bibitem{wang2023rodin}
Wang, T., Zhang, B., Zhang, T., Gu, S., Bao, J., Baltrusaitis, T., Shen, J., Chen, D., Wen, F., Chen, Q., et~al.: Rodin: A generative model for sculpting 3d digital avatars using diffusion. In: Proceedings of the IEEE/CVF Conference on Computer Vision and Pattern Recognition. pp. 4563--4573 (2023)

\bibitem{Wang2023ProlificDreamerHA}
Wang, Z., Lu, C., Wang, Y., Bao, F., Li, C., Su, H., Zhu, J.: Prolificdreamer: High-fidelity and diverse text-to-3d generation with variational score distillation. ArXiv  \textbf{abs/2305.16213} (2023), \url{https://api.semanticscholar.org/CorpusID:258887357}

\bibitem{Xu2022Dream3DZT}
Xu, J., Wang, X., Cheng, W., Cao, Y.P., Shan, Y., Qie, X., Gao, S.: Dream3d: Zero-shot text-to-3d synthesis using 3d shape prior and text-to-image diffusion models. 2023 IEEE/CVF Conference on Computer Vision and Pattern Recognition (CVPR) pp. 20908--20918 (2022), \url{https://api.semanticscholar.org/CorpusID:255340806}

\bibitem{xu2023matlaber}
Xu, X., Lyu, Z., Pan, X., Dai, B.: Matlaber: Material-aware text-to-3d via latent brdf auto-encoder. arXiv preprint arXiv:2308.09278  (2023)

\bibitem{Yang2019PointFlow3P}
Yang, G., Huang, X., Hao, Z., Liu, M.Y., Belongie, S.J., Hariharan, B.: Pointflow: 3d point cloud generation with continuous normalizing flows. 2019 IEEE/CVF International Conference on Computer Vision (ICCV) pp. 4540--4549 (2019), \url{https://api.semanticscholar.org/CorpusID:195750453}

\bibitem{Zeng2022LIONLP}
Zeng, X., Vahdat, A., Williams, F., Gojcic, Z., Litany, O., Fidler, S., Kreis, K.: Lion: Latent point diffusion models for 3d shape generation. ArXiv  \textbf{abs/2210.06978} (2022), \url{https://api.semanticscholar.org/CorpusID:252872881}

\bibitem{Zhang20223DILGIL}
Zhang, B., Nie{\ss}ner, M., Wonka, P.: 3dilg: Irregular latent grids for 3d generative modeling. ArXiv  \textbf{abs/2205.13914} (2022), \url{https://api.semanticscholar.org/CorpusID:249152155}

\bibitem{zhang20233dshape2vecset}
Zhang, B., Tang, J., Niessner, M., Wonka, P.: 3dshape2vecset: A 3d shape representation for neural fields and generative diffusion models. arXiv preprint arXiv:2301.11445  (2023)

\bibitem{Zheng2022SDFStyleGANIS}
Zheng, X., Liu, Y., Wang, P.S., Tong, X.: Sdf‐stylegan: Implicit sdf‐based stylegan for 3d shape generation. Computer Graphics Forum  \textbf{41} (2022), \url{https://api.semanticscholar.org/CorpusID:250048592}

\bibitem{Zheng2023LocallyAS}
Zheng, X., Pan, H., Wang, P.S., Tong, X., Liu, Y., yeung Shum, H.: Locally attentional sdf diffusion for controllable 3d shape generation. ACM Transactions on Graphics (TOG)  \textbf{42},  1 -- 13 (2023), \url{https://api.semanticscholar.org/CorpusID:258557967}

\bibitem{zheng2022sdf}
Zheng, X., Liu, Y., Wang, P., Tong, X.: Sdf-stylegan: Implicit sdf-based stylegan for 3d shape generation. In: Computer Graphics Forum. vol.~41, pp. 52--63. Wiley Online Library (2022)

\bibitem{Zhou20213DSG}
Zhou, L., Du, Y., Wu, J.: 3d shape generation and completion through point-voxel diffusion. 2021 IEEE/CVF International Conference on Computer Vision (ICCV) pp. 5806--5815 (2021), \url{https://api.semanticscholar.org/CorpusID:233182041}

\end{thebibliography}

\appendix
\clearpage
\setcounter{page}{1}

\section{Network Architectures and Training Details}

In this section, we explain the detailed architecture and training procedure of our mesh autoencoder, position DDPM, and feature DDPM.
We will release the code to facilitate reproducibility if the paper is accepted.

\begin{figure}[h]
\centering
\includegraphics[width=0.9\linewidth]{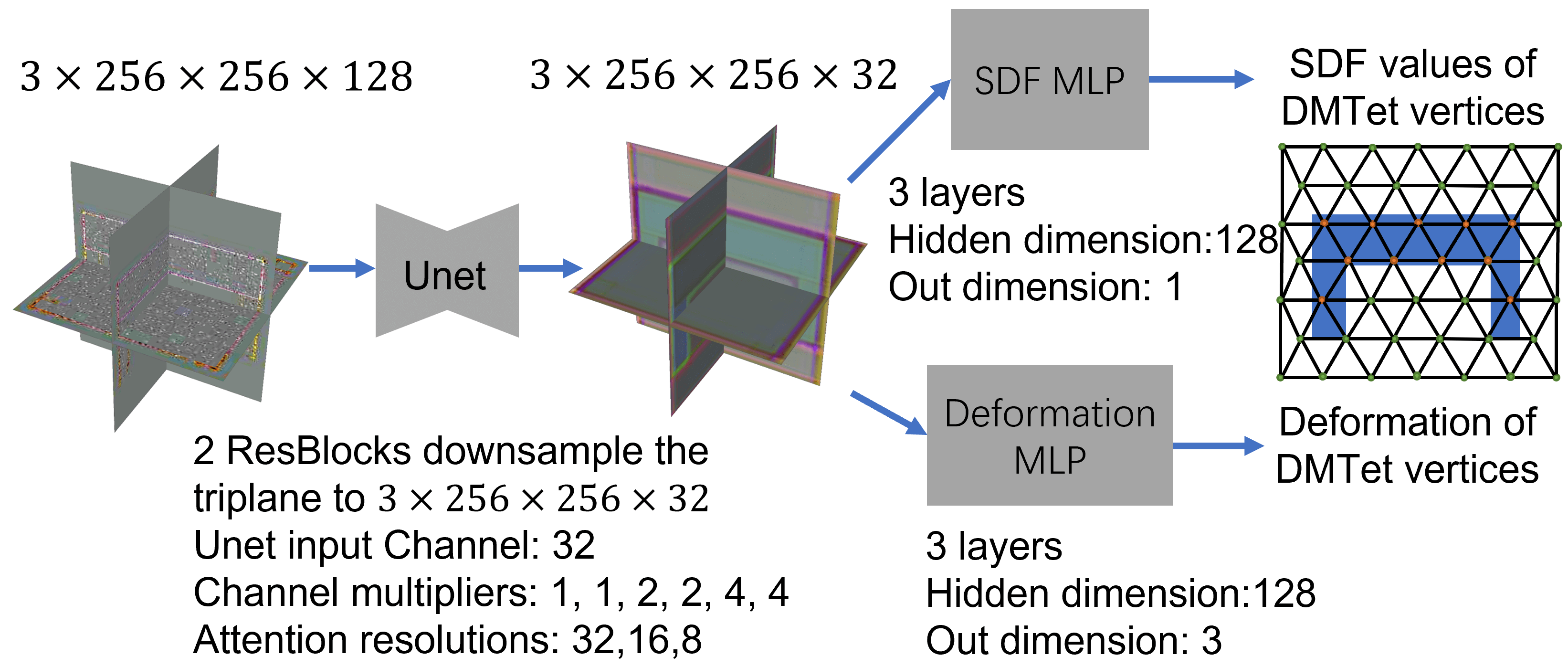}
\caption{Detailed architecture of the triplane-based decoder.}
\label{fig: mesh_decoder_with_parameters}
\end{figure}

\subsection{Mesh Autoencoder}
\label{sec: appendix_mesh_autoencoder_architecture}

\begin{figure}[t]
\centering
\includegraphics[width=0.9\linewidth]{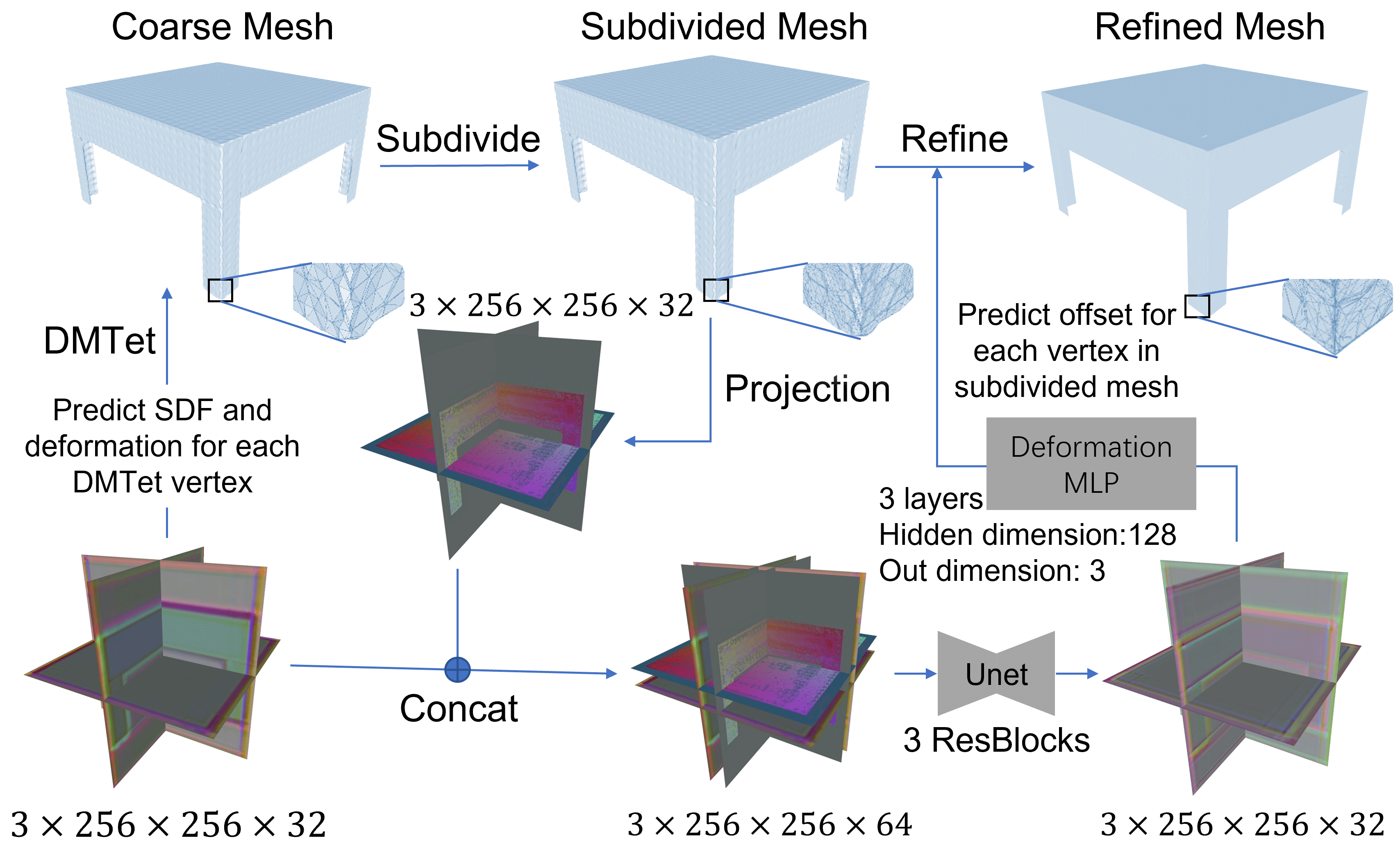}
\caption{Detailed architecture of the refinement module.}
\label{fig: mesh_refine_module_with_parameters}
\end{figure}

The mesh autoencoder is composed of the point-based encoder, the triplane-based decoder, and the refinement module.
The point-based encoder is similar to the one proposed in ~\cite{lyu2023controllable}. 
It gradually downsamples the input point cloud and propagates features level by level.
The detailed architecture of the point-based encoder is shown in Table~\ref{table: mesh_ae_architecture}.
Recall that we need to re-organize the latent point representation to the triplane representation.
We first upsample the latent points and then project them to the triplane.
The details of the upsampling process are shown in Table~\ref{table: mesh_ae_architecture}. 
The detailed architecture of the triplane-based decoder is shown in Figure~\ref{fig: mesh_decoder_with_parameters}, and the detailed architecture of the refinement module is shown in Figure~\ref{fig: mesh_refine_module_with_parameters}.

To handle varying numbers of latent points $N \in [N_{\text{min}}, N_{\text{max}}]$ in the mesh autoencoder, we pad the latent points to $N_{\text{max}}$ points with dummy points at 3D coordinate $(4,4,4)$, which are far away from the object bounding box $[-1,1]^3$, such that the dummy points will not affect features propagated to the real latent points.
The upsampling process is based on the one proposed in~\cite{lyu2023controllable}. 
In the upsampling process, each real latent point is split into $\gamma$ new points by using its feature to predict $\gamma$ offsets through a shared MLP.
The resulting points that range from $\gamma N_{\text{min}}$ to $\gamma N_{\text{max}}$ points are downsampled to a fixed number of points by furthest point sampling.
We use $\gamma=16$ in all experiments.

\begin{table*}[tb]
\centering
\caption{The schedule of $L_{\text{SDF}}$ and $L_{\text{Render}}$ to train the mesh autoencoder.}
\label{table: mesh_ae_training_schedule}
\begin{tabular}{l|cccccccccccc}
\toprule[1.5pt]
Epoch & 300 & 320 & 325 & 330 & 335 & 340 & 345 & 350 & 355 & 360 & 365 & 600 \\ \midrule[1pt]
$L_{\text{SDF}}$ & 1 & 0.9 & 0.8 & 0.7 & 0.6 & 0.5 & 0.4 & 0.3 & 0.2 & 0.1 & 0 & 0\\
$L_{\text{Render}}$ & 0 & 0.001 & 0.002 & 0.005 & 0.01 & 0.02 & 0.05 & 0.1 & 0.2 & 0.5 & 1 & 1\\
\bottomrule[1.5pt]
\end{tabular}
\end{table*}
\noindent \textbf{Training details.}
The supervision of the autoencoder is added on the latent feature $\vy$, the upsampled points $\vx_{\text{u}}$, the reconstructed mesh $\mM'$, and the predicted SDF values of the DMTet grid.
For $\vy$, we add a Kullback-Leibler divergence loss, $L_{\text{KL}}$, between $\vy$ and the standard Gaussian distribution in order to make the distribution of latent features relatively simple and smooth.
For $\vx_{\text{u}}$, we first downsample the input point cloud $\va$ to the same number of points as $\vx_{\text{u}}$ using FPS, and then add a Chamfer distance (CD), $L_{\text{CD}}$, between the downsampled points and $\vx_{\text{u}}$.
For $\mM'$, we add a rendering-based loss, $L_{\text{Render}}$, similar to the one used in~\cite{gupta20233dgen}.
Specifically, $\mM'$ is fed to a differentiable renderer to obtain the mask silhouette $\vm$ and depth map $\vd$, and the rendering-based loss is computed as the sum of L2 distance between $\vm$ and ground-truth mask silhouette, and L1 distance between $\vd$ and ground-truth depth map, averaged across $N_{\text{view}}$ views.
For the predicted SDF values of the DMTet grid, we add an MSE loss, $L_{\text{SDF}}$, between the predicted SDF values and ground-truth SDF values of the DMTet grid.

The autoencoder is trained in three phases.
In all phases, we use a weight of $10^{-7}$ for $L_{\text{KL}}$ and use a batchsize of $128$.
In the first phase,
we use a weight of $1$ for $L_{\text{CD}}$, and the other loss terms are not included.
The triplane-based decoder is not trained in this phase.
We use the Adam optimizer with a learning rate of $10^{-3}$ and train the mesh autoencoder for $300$ epochs.
The checkpoint with the lowest $L_{\text{CD}}$ is selected to train the mesh autoencoder in the second phase.

In the second phase,
$L_{\text{CD}}$ is kept with weight $1$.
We add a warm-up schedule for $L_{\text{Render}}$ and $L_{\text{SDF}}$ to ensure that DMTet can extract meaningful meshes at the initial phase.
The detailed schedule is shown in Table~\ref{table: mesh_ae_training_schedule}.
In the second phase, we do not use the refinement module and $L_{\text{Render}}$ is applied to the coarse mesh extracted from DMTet.
We use the Adam optimizer with a learning rate of $10^{-3}$ and train the mesh autoencoder for $300$ epochs.
The checkpoint with the lowest $L_{\text{Render}}$ is selected to train the mesh autoencoder in the third phase.

In the third phase, we use $L_{\text{CD}}$ with weight $1$ and $L_{\text{Render}}$ with weight $1$.
The refinement module is included and $L_{\text{Render}}$ is applied to the refined mesh.
We use the Adam optimizer with a learning rate of $5 \times 10^{-4}$ and train the mesh autoencoder for $300$ epochs.
The checkpoint with the lowest $L_{\text{Render}}$ is selected to train the latent diffusion models.
It takes about a week to train the mesh autoencoder on $32$ NVIDIA A100 GPUs in the three phases.

\begin{algorithm}[tb]
\caption{Guided sampling for position DDPM.}
\label{alg: position_ddpm_guided_sampling}
\KwIn{Trained position DDPM $\vepsilon_{\text{position}}$. Edited latent point positions $\vx_{e} \in \mathbb{R}^{N_{e} \times 3}$. Guidance scale $s$.
}
\KwOut{Sampled latent points $\vx^0$.}

Sample $\vx^T$ from $\gN(\vx^T; 0, \mI)$

\For{$t$ from $T$ to 1}{
    $\tilde{\vx}^0 =  \frac{\vx^t}{\sqrt{\bar{\alpha}_{t}}}-\frac{\sqrt{1-\bar{\alpha}_{t}} \vepsilon_{\text{position}}(\vx^t, t)}{\sqrt{\bar{\alpha}_{t}}}$

    $\mathcal{L} = \lVert \tilde{\vx}^0 [0:N_{e},:] - \vx_{e}\rVert^2$

    $\tilde{\vx}^0 \gets \tilde{\vx}^0 - s\nabla_{\tilde{\vx}^0} \mathcal{L}$

    $\tilde{\vmu}_t = \frac{ \sqrt{\bar{\alpha}_{t-1}} \beta_t }{ 1-\bar{\alpha}_{t} } \tilde{\vx}^0 + \frac{ \sqrt{\alpha_{t}} (1-\bar{\alpha}_{t-1}) }{ 1-\bar{\alpha}_{t} } \vx^t$

    $\tilde{\beta}_t = \frac{ 1-\bar{\alpha}_{t-1} }{ 1-\bar{\alpha}_{t} } \beta_t$

   Sample ${\vx}^{t-1}$ from $\gN(\vx^{t-1};\tilde{\vmu}_t, \tilde{\beta}_t \mI)$
}
\Return ${\vx}^{0}$
\end{algorithm}

\subsection{Latent Diffusion Models}
\label{sec: appendix_latent_diffusion_model}
For the position DDPMs and feature DDPMs in Section~\ref{sec: latent_diffusion_model}, we use the Transformer architecture proposed for points in ~\cite{pang2022masked}.
They all share the same architecture: The Transformer is composed of $12$ Multi-Head Self-Attettion blocks.
The embedding dimension of each attention block is $512$ and the number of attention heads is $8$.
The class label is mapped to a learnable $512$-dimension embedding and appended to the input sequence.

For the position DDPM, the 3D coordinates of the latent points are mapped to $512$-dimension embeddings through a shared MLP, and the $512$-dimension embeddings are treated as both inputs and positional embeddings to the Transformer.
The Transformer outputs are fed to a shared MLP to predict the noises added to the latent point 3D coordinates.
For the feature DDPM, both 3D coordinates and features of the latent points are fed to the Transformer.
The 3D coordinates are mapped to $512$-dimension embeddings through a shared MLP and are treated as positional embeddings.
The features are mapped to $512$-dimension embeddings through another shared MLP and are treated as inputs to the Transformer.
The Transformer outputs are fed to a shared MLP to predict the noises added to the features of the latent points.

\noindent \textbf{Traing details.}
The position and feature DDPMs are trained for $4000$ epochs with batchsize $128$ and learning rate $2 \times 10^{-4}$.
The Adam optimizer is used.
The amount of computational resources and time to train them is shown in Table~\ref{table: ddpm_training_time}.

\begin{figure}[tb]
\centering
\includegraphics[width=0.7\linewidth]{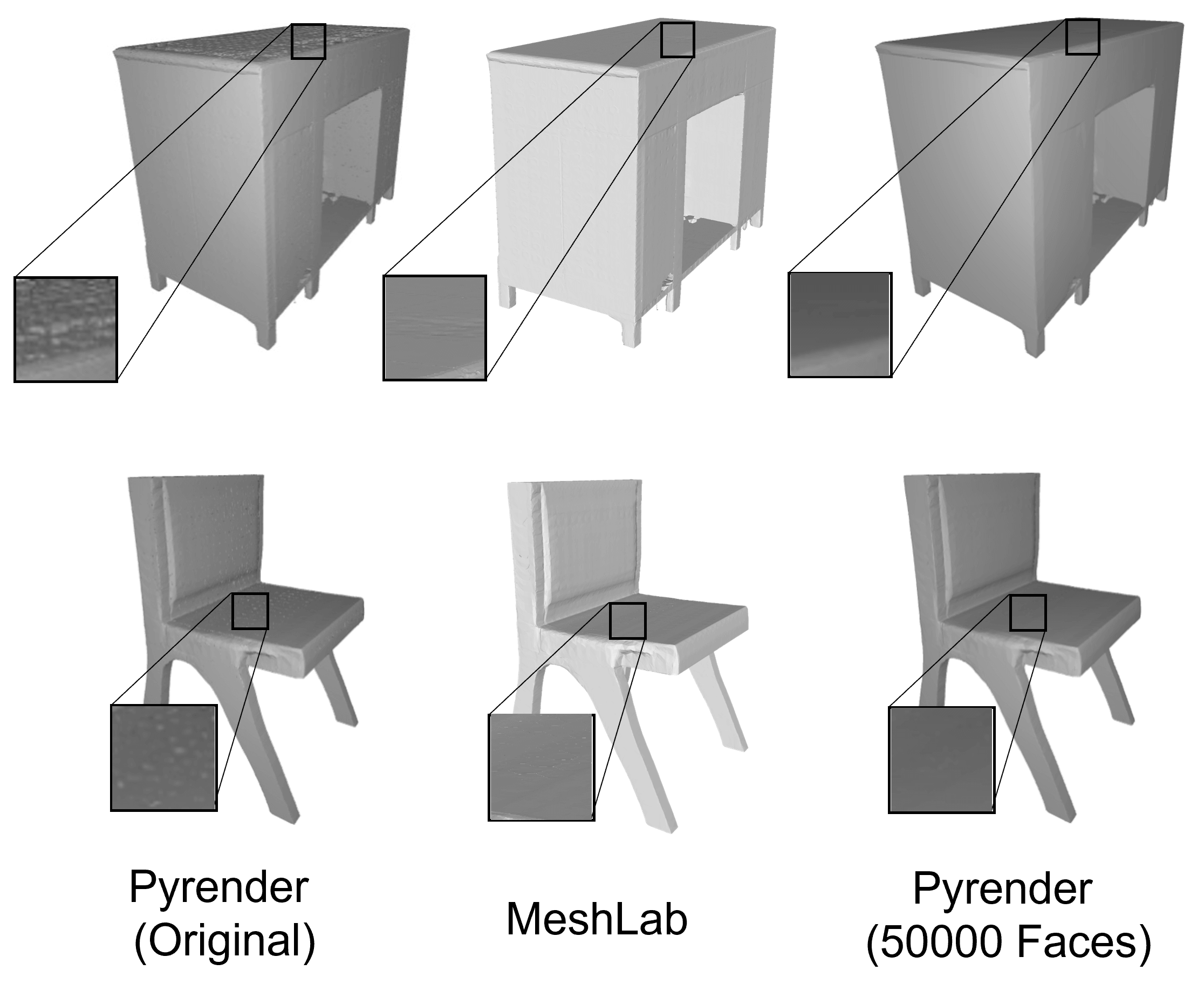}
\caption{We use Open3D's simplify-quadric-decimation to reduce the number of faces in an object to $50000$ in order to alleviate the artifacts from Pyrender.}
\label{fig: fid_5w_face_comparison}
\end{figure}

\section{Guided Sampling of Position DDPM}
\label{sec: position_ddpm_guided_sampling}

We develop a guided-sampling method for the position DDPM to facilitate controllable mesh generation and editing.
Human-edited latent points may have flaws in many cases. 
Therefore, we need to develop a method that reflects the editing effect of a user's intention, while mitigating the pitfalls of human editing.
The main idea of our method is to leverage the position DDPM's ability to generate an arbitrary number of latent points and add guidance to the sampling process of the position DDPM.

Assume human-edited latent point is $\vx_{e} \in \mathbb{R}^{N_{e} \times 3}$.
We aim to generate latent points ${\vx}^{0} \in \mathbb{R}^{N \times 3}$ that reflects changes in $\vx_{e}$, while avoids flaws in $\vx_{e}$.
We assume $N_{e} < N$. 
This can always be attained by setting $N=N_{\text{max}}$ and reducing the number of points in $\vx_{e}$ using FPS.
Next, we use $\vx_{e}$ to guide the sampling process of the position DDPM, and the detailed algorithm is shown in Algorithm~\ref{alg: position_ddpm_guided_sampling}.
We set the guidance scale $s$ to $1$ in our experiments.
The overall idea of the algorithm is to guide the first $N_{e}$ points in the $N_{\text{max}}$ generated points and encourage them to be close to the human-edited latent points.
The rest points in the $N_{\text{max}}$ generated points can fill the holes or missing parts by leveraging the generative prior in the position DDPM. 

\section{Evaluation Details}
\label{sec: evaluation_details}

\begin{figure*}[tb]
\centering
\includegraphics[width=1\linewidth]{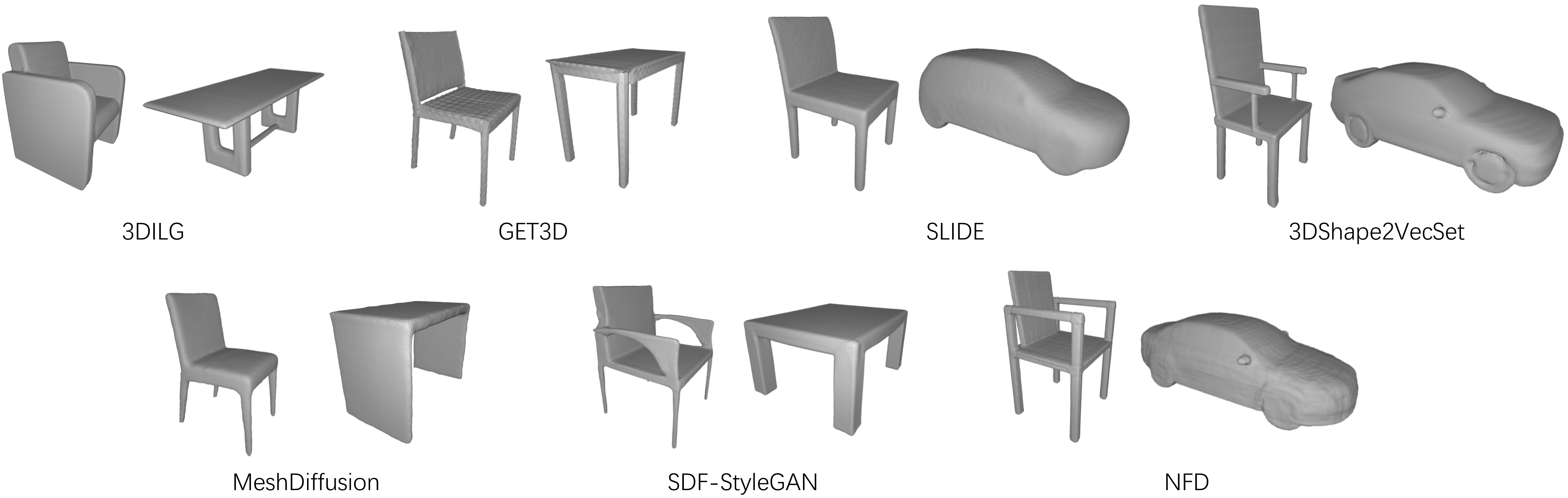}
\caption{Pyrender-rendered images of meshes generated by baseline methods.}
\label{fig: pyrender_baseline}
\end{figure*}

\noindent \textbf{Shading-FID.}
We follow \cite{zheng2022sdf} to calculate Shading-FID between rendered images of the generated meshes and meshes in the dataset.
When generating meshes using our method, we do not use classifier-free guidance for the class condition.
We use the released checkpoints of the baselines to generate meshes for evaluation.
Since the baselines use different portions of the dataset for training\footnote{They all use $70\%$ or more data for training. Therefore, the comparison with our method, which uses $70\%$ data for training, is fair.},
we use all data in a category including the training, validation, and test sets to render images as the reference set.
We generate the same number of meshes as the reference set for evaluation.
We follow the same procedures in \cite{zheng2022sdf} to render multi-view images of the generated meshes and meshes in the reference set, except that we render $512 \times 512$ images instead of $299 \times 299$ images to better capture details of the generated meshes and meshes in the reference set.

\cite{zheng2022sdf} uses Pyrender to render images by default, 
but we find that Pyrender-rendered images of the meshes generated by our method often bear obvious artifacts: There are many small dots on the mesh surfaces as shown in Figure~\ref{fig: fid_5w_face_comparison}.
We validate that this is a problem caused by Pyrender by rendering the same meshes using MeshLab.
We can see that images rendered by MeshLab do not bear any artifacts.
Therefore, it is not because meshes generated by our method have any flaws.
We hypothesize that this is because meshes generated by our method have too dense faces on their surfaces due to the face subdivision operation in the mesh autoencoder, and the vertex deformation operation in the refinement module could also result in extremely small faces.
Pyrender can not render meshes with dense and small faces properly.
We use Open3D to reduce the number of faces of the generated meshes to $50000$, and use Pyrender to render the simplified mesh again.
The result is shown in Figure~\ref{fig: fid_5w_face_comparison}.
Indeed, we can see that the artifacts are mitigated by reducing the number of faces.
Therefore, we simplify our meshes to $50000$ faces before calculating Shading-FID.

We also observe Pyrender-rendered images of meshes generated by baseline methods and find that they do not have this problem since their meshes typically have much fewer faces than our method,
and they use either Marching Cube or DMTet to extract meshes, where extremely small faces are rare.
The render result is shown in Figure~\ref{fig: pyrender_baseline}.
Therefore, the comparison between our method and baselines is fair.

\begin{table*}[tb]
\centering
\caption{Architecture of the point-based encoders and upsampling modules in the mesh autoencoders.}
\label{table: mesh_ae_architecture}
\scalebox{0.85}{
\begin{tabular}{cccccccc}
\hline
\hline
&  Input Points & Level 1 & Level 2 & Level 3 & Level 4 & Latent Points & Upsampled Points \\
\hline
Number of Points & 16384 & 4096 & 1024 & 256 & 64 & 32-64 & 256\\
Feature Dimension& 3 & 128 & 256 & 512 & 512 & 768 & 384\\
\hline
Number of Points & 16384 & 4096 & 1024 & 256 & -& 128-256 & 1024 \\
Feature Dimension& 3 & 128 & 256 & 512 & -& 192 & 192 \\
\hline
Number of Points & 16384 & 4096 & 1024 &  -& -& 512-1024 & 4096 \\
Feature Dimension& 3 & 128 & 256 & -& -& 48 & 128 \\
\hline
Number of Points & 16384 & 8192 & 4096 &  -& -& 2048-4096 & 16384 \\
Feature Dimension& 3 & 128 & 256 & -& -& 12 & 3 \\ 
\hline
\hline
\end{tabular}
}
\end{table*}
\begin{table}[tb]
\centering
\caption{Reconstruction performance of the $4$ mesh autoencoders. $N_{\text{min}}$ and $N_{\text{max}}$ are the minimum and maximum number of latent points in the latent space. $D$ is the dimension of the features of the latent points.}
\label{table: ae_reconstruct_eval}
\scalebox{1}{
\begin{tabular}{cc|ccc}
    \toprule[1.5pt]
    $N_{\text{min}}$&  $N_{\text{max}}$& $D$ & CD $\times 10^{-3}$ $\downarrow$& Mask-IoU $\uparrow$\\
    \midrule[1pt]
    32 & 64 & 768 & 1.55 & 0.954 \\
    128 & 256 & 192 & 1.34 & 0.968 \\
    512 & 1024 & 48 & \textbf{1.07} & \textbf{0.971} \\
    2048 & 4096 & 12 & 1.19 & 0.968 \\
    \bottomrule[1.5pt]
\end{tabular}
}
\end{table}

\noindent \textbf{1-NNA, MMD, Coverage.}
We follow \cite{Yang2019PointFlow3P} to compute 1-NNA, MMD, and Coverage between generated meshes and meshes in the dataset.
For each category, we randomly sample $1000$ meshes from the whole category as the reference set, and generate $1000$ meshes for our method and baselines.
We sample $2048$ points from the mesh surfaces and normalize them to the bounding box $[-1, 1]^3$ to compute 1-NNA, MMD, and Coverage.

Meshes in the ShapeNet dataset are non-watertight and have inner structures in general.
To sample points only from the mesh outer surface, we render multi-view depth maps for a mesh ($100$ random views), project each depth map to a point cloud using camera extrinsics and intrinsics, and concatenate them together to form a complete point cloud.
To obtain a uniform point cloud, we first randomly downsample the concatenated point cloud to $16384$ points and then downsample it to $2048$ points using FPS.
We use the same method to sample $2048$ points from the generated meshes.

\section{Ablation Study}
\label{sec: ablation_study}
\begin{table}[tb]
\centering
\caption{The amount of computational resources (A100 GPUs) and time to train the position DDPMs and feature DDPMs. Note that the training time could be affected by many factors such as data I/O speed, GPU utilization, or node differences in a cluster.}
\label{table: ddpm_training_time}
\scalebox{1}{
\begin{tabular}{cccc|cc}
    \toprule[1.5pt]
    \multirow{2}{*}{$N_{\text{min}}$} & \multirow{2}{*}{$N_{\text{max}}$} & \multicolumn{2}{c|}{Position DDPM} & \multicolumn{2}{c}{Feature DDPM} \\
    \cline{3-6}
     & & Number of GPUs & Days & Number of GPUs & Days \\
    \midrule[1pt]
    32 & 64 & 2 & 2 & 4 & 5\\
    128 & 256 & 4 & 2 & 16 & 3\\
    512 & 1024 & 8 & 3 & 16 & 4\\
    2048 & 4096 & 32 & 8 & 32 & 12\\
    \bottomrule[1.5pt]
\end{tabular}
}
\end{table}
\begin{table}[tb]
\centering
\caption{Ablation study of the number of latent points in the latent space. 1-NNA is reported.}
\label{table: ablation_1_nna}
\scalebox{1}{
\begin{tabular}{cc|cccc|cccc}
    \toprule[1.5pt]
    \multirow{2}{*}{$N_{\text{min}}$} & \multirow{2}{*}{$N_{\text{max}}$} & \multicolumn{4}{c|}{1-NNA (CD) $\downarrow$} & \multicolumn{4}{c}{1-NNA (EMD) $\downarrow$}\\
    \cline{3-10}
    & & Airplane & Car & Chair & Table & Airplane & Car & Chair & Table\\
    \midrule[1pt]
    32 & 64 & 95.15 & 83.45 & 80.53 & 79.63 & 94.90 & 90.00 & 83.68 & 85.19\\
    128 & 256 & \textbf{70.70} & \textbf{85.35} & \textbf{53.00} & \textbf{53.90} & \textbf{68.80} & \textbf{82.50} & \textbf{52.85} & \textbf{54.35} \\
    512 & 1024 &76.30 & 86.60 & 74.72 & 75.53 & 74.85 & 84.60 & 76.03 & 75.23 \\
    2048 & 4096 & 99.90 & 99.75 & 99.30 & 99.05 & 99.90 & 99.70 & 99.34 & 98.59 \\
    \bottomrule[1.5pt]
\end{tabular}
}
\end{table}
\begin{table}[tb]
\centering
\caption{Ablation study of the number of latent points in the latent space. MMD is reported.}
\label{table: ablation_mmd}
\scalebox{1}{
\begin{tabular}{cc|cccc|cccc}
    \toprule[1.5pt]
    \multirow{2}{*}{$N_{\text{min}}$} & \multirow{2}{*}{$N_{\text{max}}$} & \multicolumn{4}{c|}{MMD (CD) $\times 1000$ $\downarrow$} & \multicolumn{4}{c}{MMD (EMD) $\times 100$ $\downarrow$}\\
    \cline{3-10}
    & & Airplane & Car & Chair & Table & Airplane & Car & Chair & Table\\
    \midrule[1pt]
    32 & 64 & 6.52 & 5.70 & 17.84 & 21.39 & 11.99 & 9.76 & 18.49 & 18.70 \\
    128 & 256 & \textbf{3.50} & \textbf{3.99} & \textbf{14.8} & \textbf{18.53} & \textbf{8.31} & \textbf{7.93} & \textbf{15.73} & \textbf{15.45} \\
    512 & 1024 & 3.75 & 4.46 & 17.83 & 23.53 & 8.75 & 8.26 & 17.83 & 17.99 \\
    2048 & 4096 & 44.47 & 35.83 & 150.23 & 162.67 & 30.31 & 23.69 & 53.29 & 54.85 \\
    \bottomrule[1.5pt]
\end{tabular}
}

\end{table}
\begin{table}[tb]
\centering
\caption{Ablation study of the number of latent points in the latent space. Coverage is reported.}
\label{table: ablation_cov}
\scalebox{1}{
\begin{tabular}{cc|cccc|cccc}
    \toprule[1.5pt]
    \multirow{2}{*}{$N_{\text{min}}$} & \multirow{2}{*}{$N_{\text{max}}$} & \multicolumn{4}{c|}{COV (CD) (Percent) $\uparrow$} & \multicolumn{4}{c}{COV (EMD) (Percent) $\uparrow$}\\
    \cline{3-10}
    & & Airplane & Car & Chair & Table & Airplane & Car & Chair & Table\\
    \midrule[1pt]
    32 & 64 & 32.0 & 25.35 & 43.79 & 44.29 & 33.9 & 25.95 & 43.39 & 42.39 \\
    128 & 256 & \textbf{47.6} & \textbf{28.6} & \textbf{49.45} & \textbf{50.65} & \textbf{49.0} & \textbf{31.1} & \textbf{48.95} & \textbf{48.55} \\
    512 & 1024 & 39.8 & 24.7 & 29.33 & 30.43 & 41.4 & 26.0 & 28.43 & 27.83 \\
    2048 & 4096 & 2.4 & 1.4 & 2.8 & 3.4 & 2.4 & 1.8 & 2.6 & 3.0 \\
    \bottomrule[1.5pt]
\end{tabular}
}
\end{table}
\begin{table}[tb]
\centering
\caption{Average generation time (in seconds) per sample of our models tested on a single NVIDIA A100 GPU. The samples are generated in $1000$ steps without any accelerations. Note that we have included the time to reconstruct meshes from latent points with features to feature DDPM's generation time.}
\label{table: ddpm_generation_time}
\scalebox{1}{
\begin{tabular}{ccccc}
    \toprule[1.5pt]
    $N_{\text{min}}$ & $N_{\text{max}}$ & Position DDPM & Feature DDPM & Total\\
    \midrule[1pt]
    32 & 64 & 0.12 & 0.22 & 0.34\\
    128 & 256 & 0.54 & 0.63 & 1.17\\
    512 & 1024 & 4.03 & 4.11 & 8.14\\
    2048 & 4096 & 45.21 & 46.97 & 92.18\\
    \bottomrule[1.5pt]
\end{tabular}
}
\end{table}

We conduct an ablation study on the number of latent points in the latent space.
We train $4$ mesh autoencoders with different numbers of latent points $N_{\text{min}}, N_{\text{max}}$, and feature dimension $D$.
We keep $N_{\text{max}} \times D$ fixed to ensure that the number of bits of the latent representation does not change.
The architecture of the encoders and upsampling details are shown in Table~\ref{table: mesh_ae_architecture}.
The architecture of decoders is the same as the one described in Section~\ref{sec: appendix_mesh_autoencoder_architecture}. 
The mesh autoencoder with $N \in [128, 256]$ is trained for scratch as described in Section~\ref{sec: appendix_mesh_autoencoder_architecture}.
The other $3$ autoencoders are initialized by using parameters from the same modules in the autoencoder with $N \in [128, 256]$, and parameters that do not exist in the autoencoder with $N \in [128, 256]$ are randomly initialized.
The $3$ autoencoders ($N\in[32,64], N\in[512, 1024], N\in[2048, 4096]$) are trained with learning rates $5 \times 10^{-4}, 2 \times 10^{-4}, 2 \times 10^{-4}$, respectively.
They all use batchsize $128$ and are trained until convergence: $400$ epochs for $N\in[32,64], N\in[512, 1024]$, and $500$ epochs for $N\in[2048, 4096]$.
The reconstruction performance of the $4$ mesh autoencoders is shown in Table~\ref{table: ae_reconstruct_eval}.
We can see that all mesh autoencoders achieve relatively good reconstruction performance, and
the mesh autoencoder with $N \in [512, 1024]$ achieves the best reconstruction performance.

Next, we train latent diffusion models in the latent space of these autoencoders.
The DDPMs all use the same Transformer architecture described in Section~\ref{sec: appendix_latent_diffusion_model}.
The amount of computational resources and time to train the DDPMs are shown in Table~\ref{table: ddpm_training_time}.
We use Shading-FID, 1NN-A, MMD, and Coverage to evaluate their generation performance.
Shading-FID is shown in Table~\ref{table: ablation_shading_fid} in the main text, 1NN-A, MMD, and Coverage are shown in Table~\ref{table: ablation_1_nna}, Table~\ref{table: ablation_mmd}, and Table~\ref{table: ablation_cov}, respectively.
We can see that the diffusion model with $N \in [128, 256]$ achieves the best generation performance in most cases.

We also test the generation time of these latent diffusion models and the result is shown in Table~\ref{table: ddpm_generation_time}.
We can see that fewer latent points lead to faster generation.
The model with $N \in [128, 256]$ achieves a good trade-off between generation quality and speed.

\section{Controllable Shape Generation}
\label{sec: controllable_shape_generation}

\begin{figure}[tb]
\centering
\includegraphics[width=0.6\linewidth]{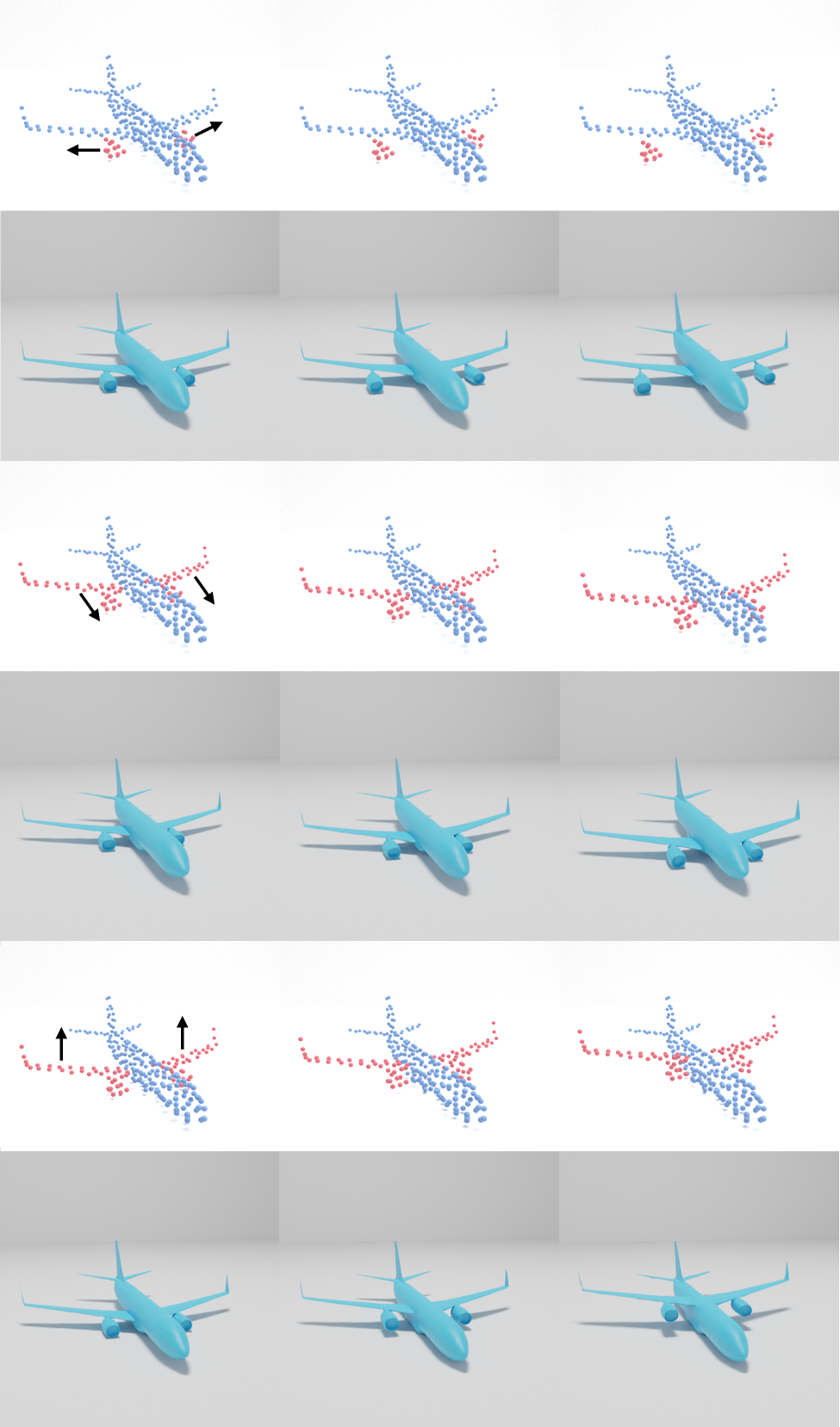}
\caption{Use positions of the latent points to control the generated shapes.
}
\label{fig: 128_256_controllable_generation}
\end{figure}

\begin{figure}[h]
\centering
\includegraphics[width=0.8\linewidth]{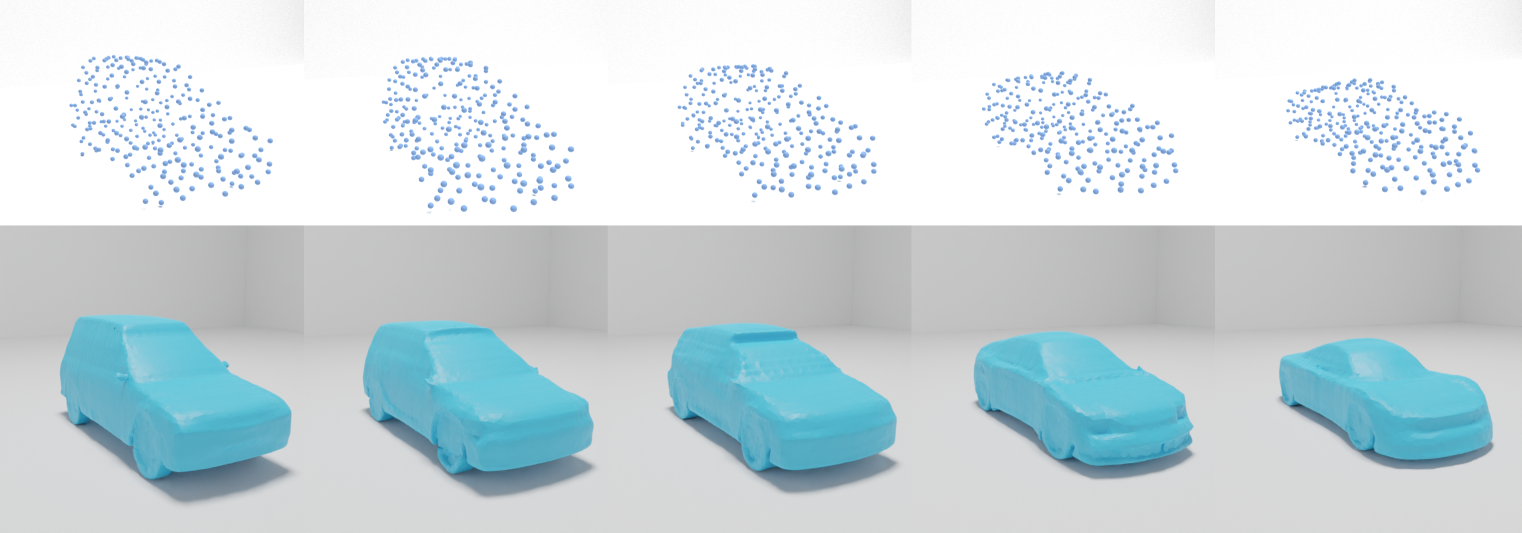}
\caption{An example of shape interpolation.
The left-most is the source shape and the right-most is the target shape.}
\label{fig: shape_interpolation_128_256_model}
\end{figure}

\begin{figure}[h]
\centering
\includegraphics[width=0.6\linewidth]{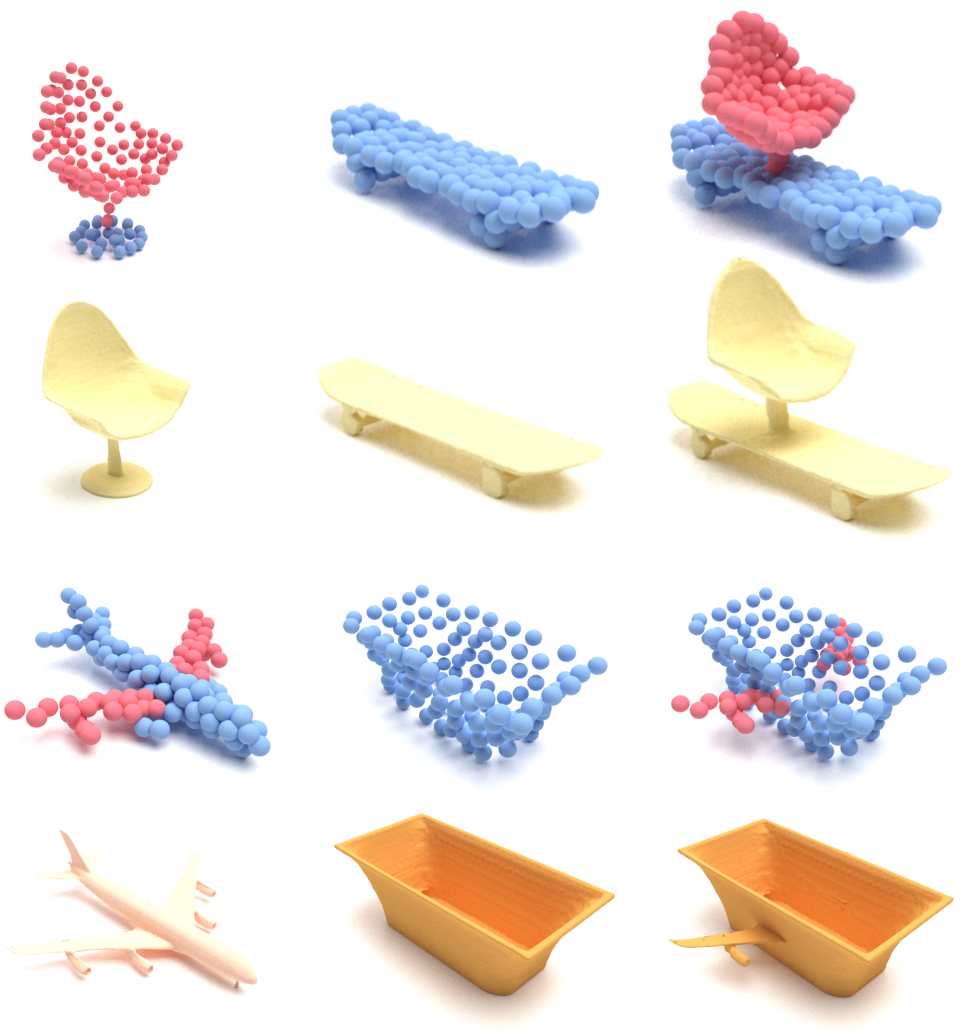}
\caption{
Examples of shape combination.
}
\label{fig: shape_combination_128_256_model}
\end{figure}

In Section~\ref{sec: controllable_mesh_generation} in the main text, we demonstrate the controllable generation ability of our method using the model with $N \in [512,1024]$.
In this section, we use our model with $N \in [128,256]$ to demonstrate controllable generation. Several examples are shown in Figure~\ref{fig: 128_256_controllable_generation}.
An example of shape interpolation is shown in Figure~\ref{fig: shape_interpolation_128_256_model}.
Examples of shape combination are shown in Figure~\ref{fig: shape_combination_128_256_model}.

\clearpage
\section{More Qualitative Results}
\label{sec: more_qualitative_results}

\begin{figure*}[h]
\centering
\includegraphics[width=1\linewidth]{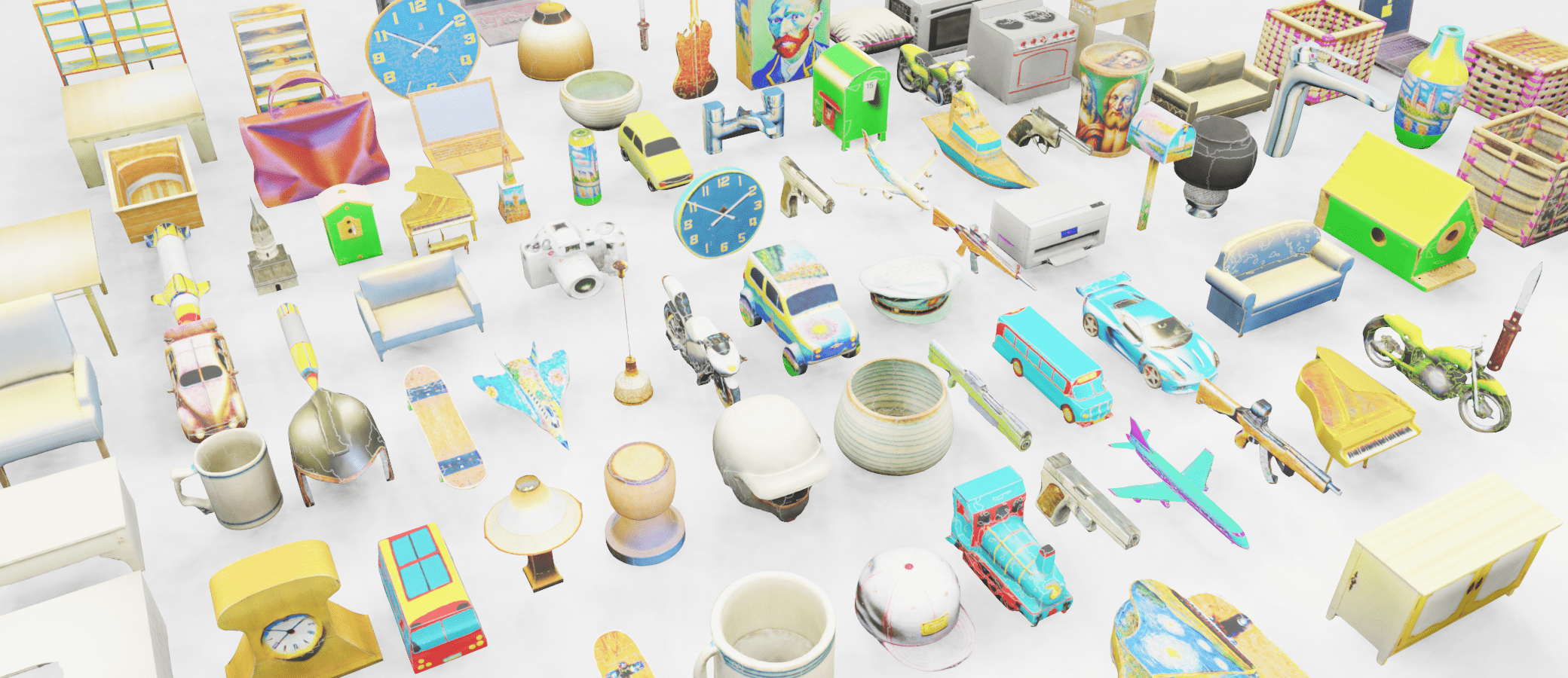}
\caption{Meshes generated by our method. Materials of the meshes are generated by an off-the-shelf material generator~\cite{xu2023matlaber}. Note that texture or material generation is NOT the focus or contribution of this work. 
We merely intend to show that our method can be seamlessly combined with off-the-shelf methods to obtain meshes with textures or materials.}
\label{fig: teaser}
\end{figure*}

\begin{figure*}[h]
\centering
\includegraphics[width=0.95\linewidth]{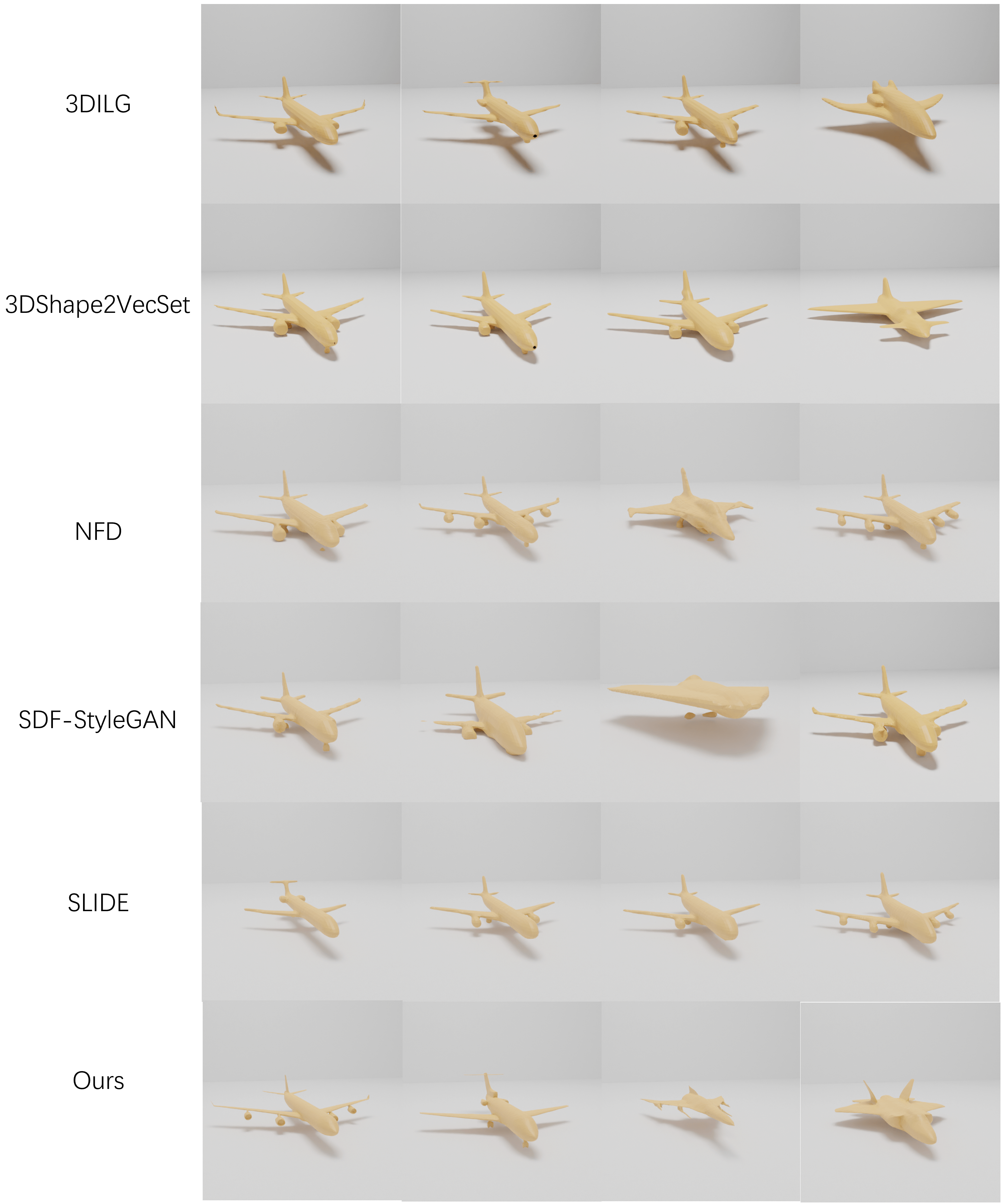}
\caption{Compare airplanes generated by our method and baselines. 
Notably, GetMesh better handles thin structures and fine details such as wings and engines than baselines.}
\label{fig: comparison_ours_and_baseline_airplane}
\end{figure*}

\begin{figure*}[h]
\centering
\includegraphics[width=0.93\linewidth]{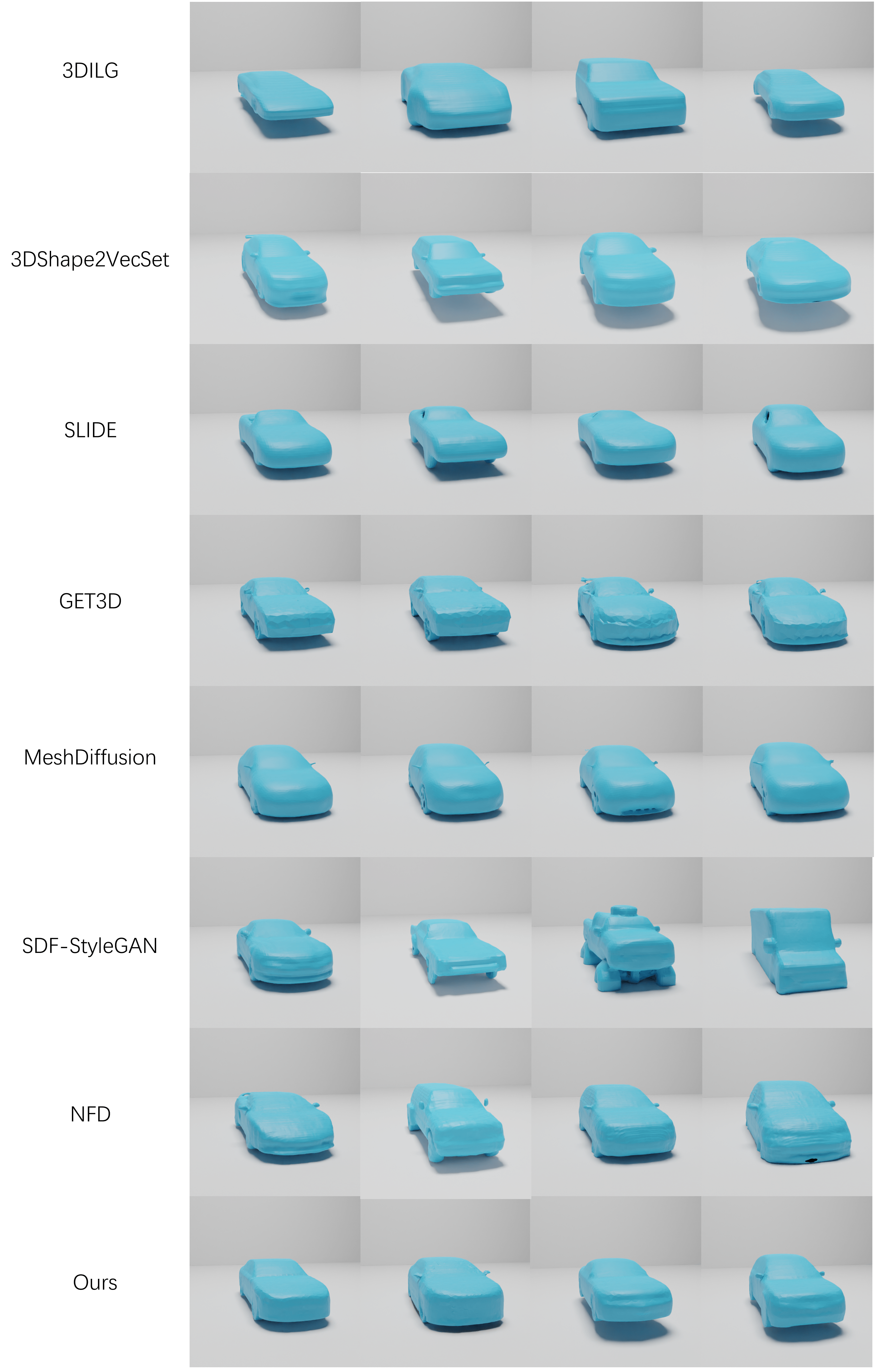}
\caption{Compare cars generated by our method and baselines. GetMesh generates meshes with smoother surfaces and sharper edges. 
}
\label{fig: comparison_ours_and_baseline_car}
\end{figure*}

\begin{figure*}[h]
\centering
\includegraphics[width=0.95\linewidth]{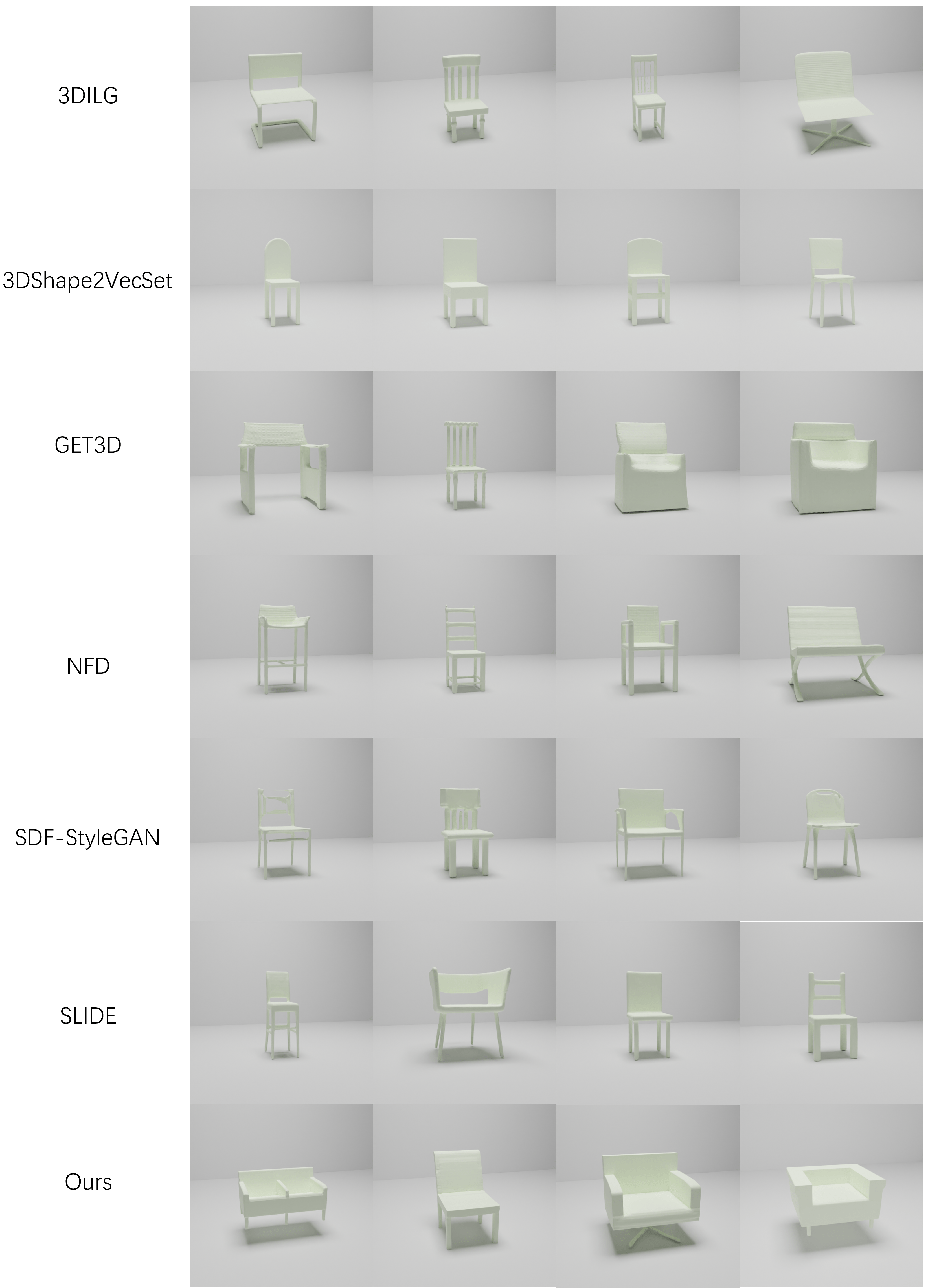}
\caption{Compare chairs generated by our method and baselines.
GetMesh generates meshes with smoother surfaces and sharper edges.}
\label{fig: comparison_ours_and_baseline_chair}
\end{figure*}

\begin{figure*}[h]
\centering
\includegraphics[width=0.95\linewidth]{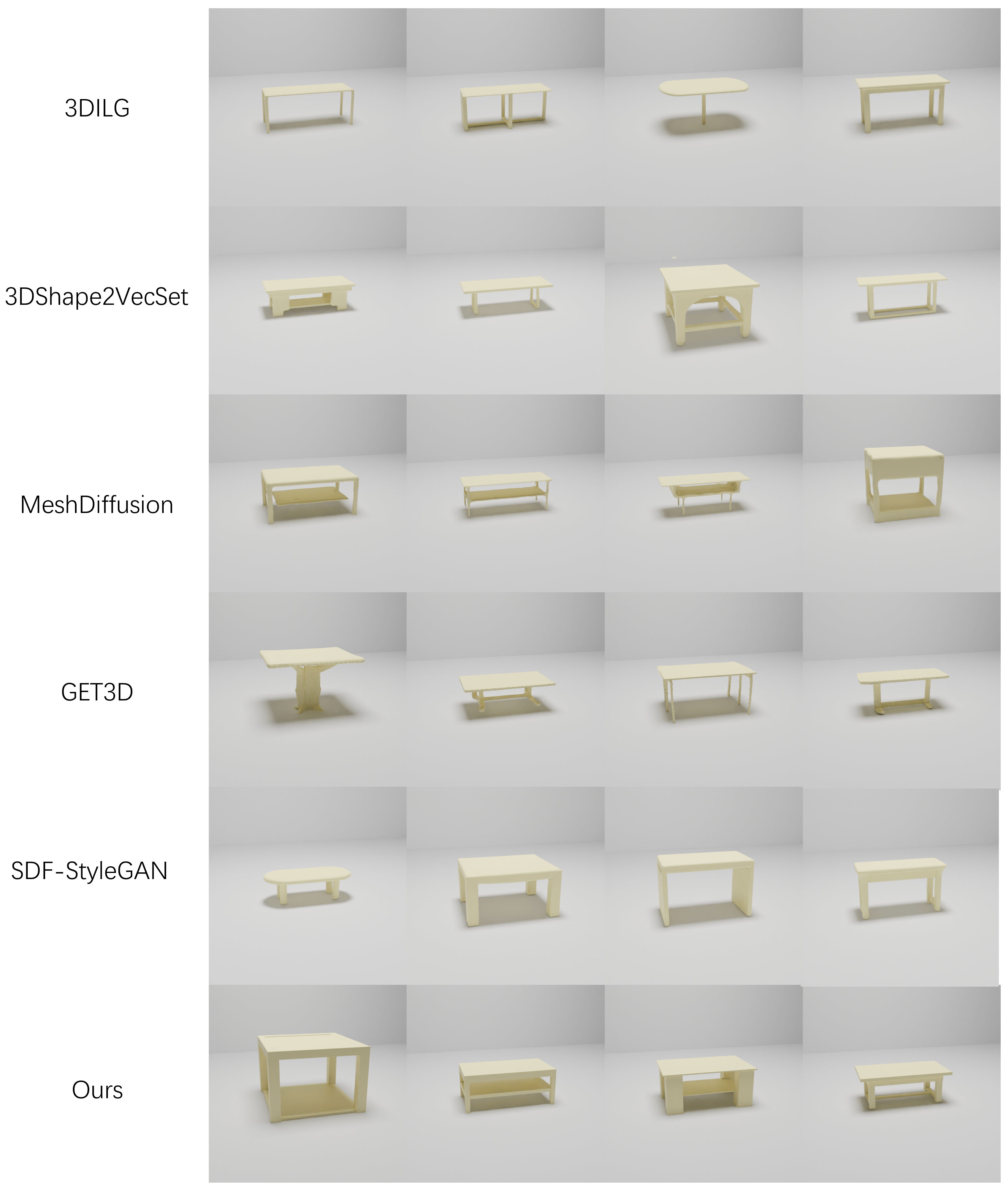}
\caption{Compare tables generated by our method and baselines.
GetMesh generates meshes with smoother surfaces and sharper edges.}
\label{fig: comparison_ours_and_baseline_table}
\end{figure*}

We generate materials for meshes in Figure~\ref{fig: teaser_no_texture} using \cite{xu2023matlaber} and result is shown in Figure~\ref{fig: teaser}.
We also qualitatively compare meshes generated by our method and baselines.
Results are shown in Figure~\ref{fig: comparison_ours_and_baseline_airplane}, Figure~\ref{fig: comparison_ours_and_baseline_car},
Figure~\ref{fig: comparison_ours_and_baseline_chair}
Figure~\ref{fig: comparison_ours_and_baseline_table}.

\end{document}